\documentclass[11pt,letterpaper]{article}

\usepackage{phew}

\newacronym{llm}{LLM}{Large Language Model}
\newacronym{rl}{RL}{Reinforcement Learning}
\newacronym{grpo}{GRPO}{Group Relative Policy Optimization}
\newacronym{cot}{CoT}{Chain-of-Thought}

\newcommand{\method}{FRPO\xspace}
\newcommand{\reg}{\mathrm{reg}}

\definecolor{lightgray}{RGB}{245,245,245}
\definecolor{darkgray}{RGB}{128,128,128}
\definecolor{purple}{RGB}{128,0,128}
\definecolor{abscol}{RGB}{250,243,224}

\titleformat{\section}[block]
{\large\bfseries}
{\thesection}{1em}{}

\newcommand{\weblink}[2]{\href{#1}{\textcolor{magenta}{#2}}}

\begin{document}

\thispagestyle{empty}

\begin{center}
\vspace*{-0.5in}
\begin{minipage}{0.3\textwidth}
\raggedleft
\includegraphics[width=\textwidth]{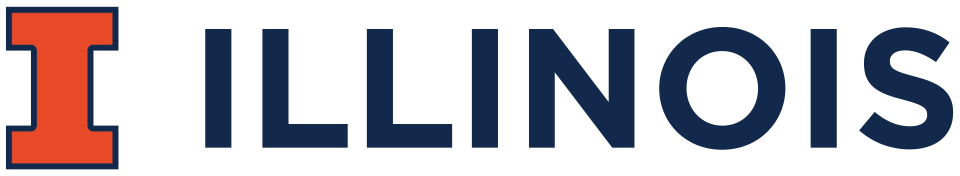} 
\end{minipage}
\end{center}

\hrule
\vspace{0.3in}

\begin{center}
{\LARGE\bfseries Future-KL Regularized GRPO: Process-Level Credit Assignment from $f$-Divergence Regularization}
\end{center}

\begin{center}
\begin{minipage}{\textwidth}
\centering
Jiarui Yao, Ruida Wang, Hao Bai, Tong Zhang

\textit{\{jiarui14,tozhang\}@illinois.edu}
\vspace{0.1in}

University of Illinois Urbana-Champaign 
\end{minipage}
\end{center}


\begin{tcolorbox}[
    colback=abscol,
    colframe=abscol,
    boxrule=0.5pt,
    arc=3pt,
    left=10pt,
    right=10pt,
    top=10pt,
    bottom=10pt
]
\gls{grpo} is widely used for critic-free \gls{llm} post-training, but its KL regularization is usually implemented as a local loss-side token penalty. We show that this misses the policy-gradient signal induced by autoregressive KL regularization. Unlike standard KL-regularized \gls{rl} objectives, GRPO's group normalization induces a non-linear prompt-level utility; for binary verifier rewards, this utility is $2\arcsin\sqrt p$. As a result, reward and KL cannot be fused before normalization without changing the implicit objective.
We derive the on-policy gradient of GRPO-style objectives with token-wise $f$-divergence regularization. The reward term recovers the standardized GRPO advantage, while the regularizer term includes a causal future-regularization return-to-go omitted by local KL losses. For reverse KL, this yields a simple future KL correction: add a reverse cumulative sum of per-token log ratios after advantage construction. The resulting method, \textbf{F}uture-KL \textbf{R}egularized \textbf{P}olicy \textbf{O}ptimization (\textbf{\method{}}), requires no critic or extra model passes. On mathematical reasoning tasks, \method{} improves pass@16 in our main large-model setting while maintaining higher entropy and lower policy drift than conventional loss-side KL baselines.

\textbf{Code:} \weblink{https://github.com/ScaleML/KL-in-LLM-RL}{https://github.com/ScaleML/KL-in-LLM-RL}
\end{tcolorbox}



\section{Introduction}

\gls{rl} has become a central component of post-training \glspl{llm} for mathematical reasoning and other verifiable tasks. A widely used paradigm is to sample multiple responses for each prompt, score them using an outcome verifier, and update the policy toward responses that outperform others in the same group. \gls{grpo} \citep{shao2024deepseekmath,guo2025deepseek} follows this paradigm and removes the need for a learned critic by normalizing outcome rewards within each prompt group. This critic-free design has made \gls{grpo} and its variants attractive for large-scale reasoning training, where outcome rewards are often sparse, binary, and inexpensive to verify.

A key ingredient in such training pipelines is KL regularization against a reference policy. KL regularization stabilizes policy updates, prevents excessive drift from the supervised model, and preserves general capabilities while optimizing for task-specific rewards. In the original GRPO formulation, the KL penalty is added as a token-level loss-side regularizer, often using a sampled estimator such as the $k3$ estimator \citep{schulman2020klapprox}. This implementation is simple and empirically useful, but it obscures several distinct questions: which sampling distribution the sampled KL term estimates under, whether its gradient corresponds to the intended KL objective, and, more importantly, how KL regularization should assign credit across tokens in an autoregressive trajectory.

\begin{figure}
    \centering
\vspace{-0.75em}
    \makebox[\textwidth][c]{
    \includegraphics[width=1.15\linewidth]{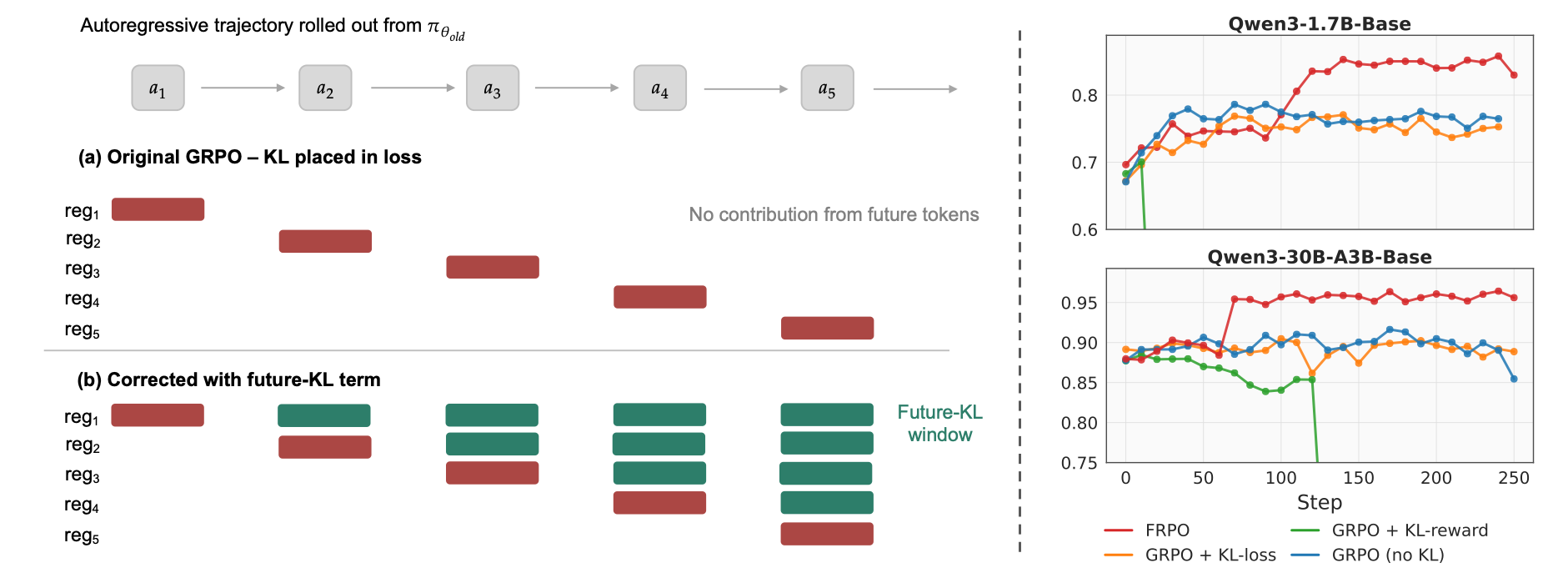}}
\vspace{-0.75em}
    \captionof{figure}{\textbf{Left:} Accumulating future KL divergence between the current policy $\pi_\theta$ and the reference policy $\pi_{\rm ref}$ contributes to the trajectory-level regularization signal, but is not captured by an independent loss-side token penalty. \textbf{Right:} The dynamics of performance pass@16 for Qwen3-1.7B-Base and Qwen3-30B-A3B-Base under different KL integrations.}
    \label{fig:teaser}
\end{figure}

Recent analyses \citep{zhang2025design} of KL-regularized policy-gradient methods have clarified several implementation-level issues, including stale-rollout distribution mismatch and the gap between value-unbiased and gradient-unbiased sampled KL estimators. These observations are important, but they do not fully answer a more basic on-policy question: what token-level policy-gradient signal is induced by regularization itself when the expectation is over the current autoregressive policy?

We therefore separate two issues that are often conflated. The first is an estimator issue: rollouts may be sampled from $\pi_{\theta_{\rm old}}$ while a KL term is evaluated under $\pi_\theta$, requiring importance correction or fresh log-probabilities. The second is an objective issue: even when samples are on-policy, a loss-side local KL term differentiates only the sampled token penalty and ignores how the current token changes the distribution of future states. This paper focuses on the second issue and shows that it leads to a different token-level credit-assignment rule.

This issue is fundamental in autoregressive language modeling. An early token not only incurs its own local regularization cost, but also changes the future prefix distribution and hence the regularization costs of later tokens. Consequently, token-wise $f$-divergence regularization induces a process-level credit assignment signal. For reverse KL, the policy-gradient weight for each token should include a future-KL return-to-go term, rather than only an independent token-wise loss penalty.

\paragraph{Why GRPO is not the standard MaxEnt-RL setting.} Existing analyses of KL-regularized policy gradients (soft-Q, MaxEnt RL, KL-in-reward PPO/GAE) target objectives that are \emph{linear} in reward, $\mathbb E_\pi[r]-\beta\,\mathbb E_\pi[\KL]$. Linearity is what allows reward and regularizer to be folded into a single augmented reward, and the future-KL return-to-go is the textbook consequence of differentiating an autoregressive expectation in this regime. GRPO is structurally different. GRPO does not use a value-function baseline; it constructs advantages by group-normalizing outcome rewards, which makes the effective prompt-level objective \emph{non-linear} in the expected reward — for binary verifier rewards, $\ell(p)=2\arcsin\sqrt{p}$ rather than the raw pass rate $p$. This non-linearity has two consequences absent from prior derivations. First, reward and KL are no longer fuseable: $\ell(\mathbb E[r-\beta\,\KL])\neq\ell(\mathbb E[r])-\beta\,\mathbb E[\KL]$, so naively adding a KL penalty into the reward before group normalization changes the group mean and standard deviation, distorts the implicit arcsin objective, and collapses token-level KL information into a single trajectory-level scalar shared by all tokens. Second, the on-policy gradient acquires a prompt-dependent gain $\ell'(p)$ on the reward term that the regularizer term does \emph{not} inherit, so the natural insertion point for a token-level future-KL correction is \emph{after} advantage construction, not via reward shaping.

We derive the on-policy gradient of a GRPO-style objective under this general non-linear-$\ell$ + token-wise-$f$ structure, and show that the resulting token-level weight decomposes into two terms: an outcome-level group-relative advantage (the score of the non-linear utility $\ell$) and a future regularization correction. This leads to a decoupled future-KL formulation for reverse KL: compute the GRPO advantage from the original verifier reward, and then add a token-dependent future-KL term without modifying the group reward statistics. This separates trajectory-level ranking from process-level KL credit assignment, and differs from loss-side KL correction, KL-in-reward normalization, and actor-critic GAE methods, which respectively address surrogate estimation, coupled reward shaping under \emph{linear} reward objectives, and value-based credit assignment, rather than critic-free GRPO under a non-linear utility with token-dependent KL regularization.

In summary, this work makes the following contributions:
\begin{enumerate}[leftmargin=2em,itemsep=0.25em,topsep=0.25em,parsep=0pt,after=\vspace{-1.1em}]
    \item We identify that GRPO's group normalization induces a \emph{non-linear} prompt-level utility $\ell$ (the arcsin transform for binary verifier rewards), and we show that this non-linearity makes reward and KL non-fuseable: KL-in-reward modifies the implicit GRPO objective in a way that vanishes only when $\ell$ is linear.
    \item Under this non-linear-$\ell$ + token-wise-$f$ setting, we derive the on-policy policy gradient. The reward term inherits a prompt-dependent gain $\ell'$ that the regularizer term does not, recovering the standardized GRPO advantage as the score of the arcsin utility, while the regularizer term decomposes into a local token penalty plus a causal future-regularization return-to-go. The classical loss-side surrogate misses this future term even on-policy.
    \item The asymmetry between $\ell'$-weighted reward and unweighted regularizer dictates a decoupled placement: we propose \method{}, which preserves GRPO's outcome advantage and adds a token-level future-KL correction \emph{after} advantage construction. Experiments demonstrate its effectiveness in our main mathematical-reasoning setting.
\end{enumerate}

\section{Preliminaries and Notations}
\label{sec:prelim}

Consider an autoregressive language model $\pi_\theta$ generating a response $o = (o_1, o_2, \ldots, o_T)$ given a prompt $x$, $\theta$ represents the model's parameters. We write the conditional probability of generating response $o$ from the prompt $x$ as
\(
    \pi_\theta(o \mid x) = \prod_{t=1}^{T} \pi_\theta(o_t \mid x, o_{<t}).
\)

Let $\pi_{\rm ref}$ denote a frozen reference policy, and $\pi_{\theta_{\rm old}}$ denote the policy at the time of rollout (i.e., the sampling policy) which is an older version of the current policy $\pi_\theta$. 

We define the following shorthand for token-level quantities at position $t$:
\begingroup
\setlength{\abovedisplayskip}{4pt}
\setlength{\belowdisplayskip}{4pt}
\setlength{\abovedisplayshortskip}{2pt}
\setlength{\belowdisplayshortskip}{2pt}
\setlength{\jot}{1pt}
\begin{align*}
    \delta_t &\triangleq \log \pi_\theta(o_t \mid x, o_{<t}) - \log \pi_{\rm ref}(o_t \mid x, o_{<t}) = \log\frac{\pi_\theta(o_t|x,o_{<t})}{\pi_{\rm ref}(o_t|x_t,o_{<t})}, \\
    \alpha_t &\triangleq \frac{\pi_\theta(o_t \mid x, o_{<t})}{\pi_{\rm ref}(o_t \mid x, o_{<t})} = e^{\delta_t}, \quad
    \rho_t\triangleq \frac{\pi_\theta(o_t \mid x, o_{<t})}{\pi_{\theta_{\rm old}}(o_t \mid x, o_{<t})}.
\end{align*}
\endgroup

For the ease of derivation, we further define $u_t\triangleq \alpha_t^{-1}=\pi_{\rm ref}/\pi_\theta$.

The true token-level reverse KL divergence at position $t$ (given context $s_t = (x, o_{<t})$) is:
\begingroup
\setlength{\abovedisplayskip}{4pt}
\setlength{\belowdisplayskip}{4pt}
\setlength{\abovedisplayshortskip}{2pt}
\setlength{\belowdisplayshortskip}{2pt}
\setlength{\jot}{1pt}
\begin{equation}\label{eq:true_kl}
    \KL\big[\pi_\theta(\cdot \mid s_t) \,\|\, \pi_{\rm ref}(\cdot \mid s_t)\big] = \sum_{a \in \mathcal{V}} \pi_\theta(a \mid s_t) \log \frac{\pi_\theta(a \mid s_t)}{\pi_{\rm ref}(a \mid s_t)},
\end{equation}
\endgroup
where $\mathcal{V}$ is the vocabulary. Here we slightly abuse the notations with both $a$ and $o$ representing the actions taken by the policy $\pi$, and denote $a^{(j)}=(a_1,a_2,\cdots,a_j)$ as the generated tokens so far, and $a^{(-j)}=(a_{j+1},a_{j+2},\cdots,a_T)$ the remaining tokens.

\noindent\textbf{KL Estimators} \citet{schulman2020klapprox} proposes three widely adopted KL estimators, $\hat D^{k1}=\log\ \alpha$, $\hat D^{k2}=(\log\alpha)^2/2$, and $\hat D^{k3}=1/\alpha+\log\alpha-1$, for estimating reverse KL $\KL(\pi_\theta\|\pi_{\rm ref})$, among which $k3$ is used most commonly due to its unbiased value and nonnegative property. We provide a thorough analysis of the expectation and gradient characteristics in Appendix~\ref{sec:app:kl_estimators}. The corresponding forward KL estimators could be defined similarly, with the ratio $\alpha$ being simply reversed. While these estimators motivate the main examples, our policy-gradient derivation in Section~\ref{sec:correct_future_kl} applies to general token-wise $f$-divergence generators.
\vspace{-1em}
\section{Classical GRPO and the Regularization Question}
\label{sec:classical_grpo}

\subsection{GRPO Objective and Group-Normalized Advantage}

DeepSeek-Math~\citep{shao2024deepseekmath} proposes \gls{grpo} as a critic-free policy optimization method. For each prompt $x$, it samples a group of $G$ responses $\{o_i\}_{i=1}^G \sim \pi_{\theta_{\rm old}}(\cdot \mid x)$, evaluates scalar outcome rewards $r_i=r(x,o_i)$, and optimizes a PPO-style surrogate:
\begingroup
\normalsize
\begin{equation}\label{eq:grpo}
    \mathcal{J}_{\rm GRPO}(\theta)
    =
    \mathbb{E}\bigg[
    \frac{1}{G}\sum_{i=1}^{G}\frac{1}{|o_i|}\sum_{t=1}^{|o_i|}
    \Big\{
    \min\!\left[\rho_{i,t}\hat A_{i,t},
    \operatorname{clip}(\rho_{i,t},1-\varepsilon,1+\varepsilon)\hat A_{i,t}\right]
    -\beta\,\hat D_{i,t}
    \Big\}
    \bigg],
\end{equation}
\endgroup
where $\rho_{i,t}=\pi_\theta(o_{i,t}\mid x,o_{i,<t})/\pi_{\theta_{\rm old}}(o_{i,t}\mid x,o_{i,<t})$ and $\hat D_{i,t}$ is a sampled token-level KL penalty against the reference policy.

Unlike PPO with a learned value function, GRPO constructs advantages by normalizing rewards within the prompt group:
\begin{align*}
    \mu_G = \frac{1}{G}\sum_{i=1}^G r_i,
    \qquad
    \sigma_G = \sqrt{\frac{1}{G}\sum_{i=1}^G (r_i-\mu_G)^2},
\end{align*}
and assigns every token in response $i$ the same outcome advantage
\begin{equation}\label{eq:grpo_adv}
    \hat A_{i,t} = \hat A_i^{\rm GRPO}
    =
    \frac{r_i-\mu_G}{\sigma_G},
    \qquad
    t=1,\ldots,|o_i|.
\end{equation}
This design removes the value model but also means that GRPO's reward signal is trajectory-level: before any KL term is added, all tokens in the same response receive the same credit.

\subsection{The Implicit Non-Linear Utility from Group Normalization}

Before discussing KL regularization, we record a consequence of group normalization that is the structural source of every novelty in this paper. The arcsin objective induced by standardized binary rewards has also been noted in REINFORCE-Ada~\citep{xiong2025reinforce}. We revisit it here not as a new policy-gradient observation, but to expose a GRPO-specific complication: in binary-reward reasoning tasks, the normalization in Eq.~\eqref{eq:grpo_adv} is not a neutral preprocessing step; it determines a \emph{non-linear} prompt-level utility followed by the population GRPO gradient. Standard derivations of KL-regularized policy gradients (soft-Q, MaxEnt RL, KL-in-reward PPO) target objectives that are \emph{linear} in reward, $\mathbb E_\pi[r]-\beta\,\mathbb E_\pi[\KL]$, where reward and regularizer fuse into a single augmented reward. GRPO breaks that linearity, and as we show below, this is what makes putting KL into the reward before normalization structurally — not just numerically — incorrect.

\begin{proposition}[Implicit objective of binary GRPO]\label{prop:grpo_arcsin_objective}
Fix a prompt $x$ with binary reward $r(x,a)\in\{0,1\}$ and pass probability $p_\theta(x)=\mathbb E_{a\sim\pi_\theta(\cdot|x)}[r(x,a)]$. Replacing finite-group statistics with their population values, the expected GRPO policy gradient for this prompt is
\begin{align*}
    \mathbb E_{a\sim\pi_\theta(\cdot|x)}
    \left[
    \frac{r(x,a)-p_\theta(x)}{\sqrt{p_\theta(x)(1-p_\theta(x))}}
    \nabla_\theta\log\pi_\theta(a|x)
    \right]
    =
    \nabla_\theta\left[2\arcsin\sqrt{p_\theta(x)}\right].
\end{align*}
\end{proposition}

The proof is given in Appendix~\ref{app:arcsin-allocation}. Group normalization should therefore be viewed as part of the objective, not merely as a numerical rescaling: the standardized GRPO advantage is the score of $\ell(p)=2\arcsin\sqrt{p}$, with $\ell'(p)=1/\sqrt{p(1-p)}$. Because $\ell$ is non-linear, the operations of folding KL into the reward and applying $\ell$ do not commute, $\ell(\mathbb E[r-\beta\,\KL])\neq \ell(\mathbb E[r])-\beta\,\mathbb E[\KL]$, and modifying the reward before normalization changes the implicit arcsin objective by an amount that depends on the per-prompt KL distribution and the curvature of $\ell$.

\begin{proposition}[KL-in-reward distorts the implicit utility under non-linear $\ell$]\label{prop:kl_in_reward_distortion}
Let $\tilde r_i=r_i-\beta\sum_{j}\delta_{i,j}$ and $\hat A_i^{\rm KLR}=(\tilde r_i-\tilde\mu_G)/\tilde\sigma_G$. In the population limit, the gradient driven by $\hat A^{\rm KLR}$ targets a prompt-level objective that reduces to $\ell(\mathbb E[r])-\beta\,\mathbb E[\KL]$ \emph{only when $\ell$ is affine}. Under GRPO's $\ell(p)=2\arcsin\sqrt p$, the discrepancy is non-vanishing whenever the group has non-trivial KL spread.
\end{proposition}
GRPO's outcome advantage should continue to rank responses by verifier reward, while KL regularization should provide a separate token-level credit-assignment signal — this is the structural reason behind the general regularized objective in Eq.~\eqref{eq:general_f_objective}, and the predicted ``KL-in-reward collapse'' is observed in Section~\ref{sec:experiment} (Appendix~\ref{sec:kl_reward_normalization} discusses the alternative placements in detail).

\subsection{Loss-Side KL in the Classical Formulation}

The KL term commonly used in GRPO implementations is the sampled $k3$ estimator~\citep{schulman2020klapprox}. With $u_t=\pi_{\rm ref}(o_t\mid s_t)/\pi_\theta(o_t\mid s_t)=\alpha_t^{-1}$, it is
\begin{equation}\label{eq:k3_form}
    \hat D_t^{(k3)}(\pi_\theta\|\pi_{\rm ref})
    =
    u_t-\log u_t-1
    =
    \frac{1}{\alpha_t}-\log\frac{1}{\alpha_t}-1.
\end{equation}

\begin{proposition}[Value unbiasedness under on-policy sampling]\label{prop:k3_unbiased}
For a fixed state $s_t$, $\hat D_t^{(k3)}$ is an unbiased value estimator of $\KL[\pi_\theta(\cdot|s_t)\|\pi_{\rm ref}(\cdot|s_t)]$ when the sampled token is drawn from $\pi_\theta(\cdot|s_t)$.
\end{proposition}

This proposition is about the scalar value of a sampled KL estimator. It does not by itself imply that placing $\hat D_t^{(k3)}$ as a local differentiable loss term gives the policy gradient of a KL-regularized autoregressive objective. Even in the on-policy case $\pi_{\theta_{\rm old}}=\pi_\theta$, a token $a_i$ changes not only its own local KL term but also the distribution of future prefixes, and therefore the KL costs at later positions. The next section derives this missing causal term directly.

\section{On-Policy f-Divergence Regularization}
\label{sec:correct_future_kl}

\subsection{General Policy Gradient}

Consider the on-policy regularized objective
\begin{equation}\label{eq:general_f_objective}
    Q_f(\theta)
    =
    \mathbb E_x\,\ell\!\left(\mathbb E_{a\sim\pi_\theta(\cdot|x)} r(x,a)\right)
    -
    \beta\,
    \mathbb E_x\mathbb E_{a\sim\pi_\theta(\cdot|x)}
    \sum_{i=1}^{T} f(u_i),
    \qquad
    u_i=\frac{\pi_{\rm ref}(a_i\mid s_i)}{\pi_\theta(a_i\mid s_i)}.
\end{equation}
Here $f$ is a differentiable convex generator applied to the token-level likelihood ratio; KL is only one special case. The function $\ell$ captures the prompt-level reward objective induced by the advantage construction. For binary-reward GRPO, Proposition~\ref{prop:grpo_arcsin_objective} gives $\ell(p)=2\arcsin\sqrt p$. Thus the derivation below applies to GRPO as a special case while also covering a broader class of token-wise $f$-divergence regularizers between the current and reference policies. The formulation should be read as a sampled autoregressive analogue of per-state $f$-divergence regularization: at each generated prefix $s_i$, the penalty is evaluated on the sampled action through the ratio $\pi_{\rm ref}(a_i|s_i)/\pi_\theta(a_i|s_i)$, and the expectation over trajectories supplies the state distribution induced by $\pi_\theta$.

\begin{theorem}[On-policy gradient with token-wise regularization]\label{thm:f_divergence_pg}
For any baseline $b(s_i)$ that does not depend on the current action $a_i$, the gradient of Eq.~\eqref{eq:general_f_objective} can be written as
\begin{align}
    \nabla_\theta Q_f(\theta)
    =
    \mathbb E_x\mathbb E_{a\sim\pi_\theta(\cdot|x)}
    \sum_{i=1}^{T}
    w_i\,
    \nabla_\theta\log\pi_\theta(a_i\mid s_i),
    \qquad
    w_i
    =
    \operatorname{sg}\!\left(
    A_i-\beta\,\reg_i
    \right),
    \label{eq:f_divergence_pg}
\end{align}
where
\begin{align}
    A_i
    &=
    \ell'\!\left(\mathbb E_{a\sim\pi_\theta} r(x,a)\right)r(x,a)-b(s_i),
    \label{eq:general_reward_advantage}
    \\
    \reg_i
    &=
    \sum_{j\ge i} f(u_j)-f'(u_i)u_i
    =
    \sum_{j>i} f(u_j)+f(u_i)-f'(u_i)u_i .
    \label{eq:kl_regularized_pg_weight}
\end{align}
\end{theorem}

Here $\operatorname{sg}$ denotes the implementation convention that the sampled coefficient is treated as a stop-gradient weight in the policy-gradient surrogate; the equality is the corresponding likelihood-ratio gradient identity. The proof is in Appendix~\ref{sec:app:proofs:kl_regularized_pg_weight}. Two structural features of Eq.~\eqref{eq:f_divergence_pg} drive every algorithmic choice that follows. First, $\ell$ multiplies $A_i$ but \emph{not} $\reg_i$: the reward signal carries a prompt-dependent gain $\ell'(\mathbb E\, r)$, while the regularizer term is independent of $\ell$. This asymmetry is invisible in linear-reward MaxEnt RL ($\ell(p)=p$, $\ell'\equiv 1$) and is what blocks the standard ``fold KL into reward'' move under GRPO normalization: any reward shaping is rescaled by $\ell'$ along with the verifier reward, whereas the genuine token-level regularizer is not. Second, the sum over $j\geq i$ is the gradient through the future state distribution induced by the autoregressive policy, and it is exactly what a purely local loss-side penalty misses by differentiating only $f(u_i)$ at the sampled token.

The theorem also clarifies why the issue is not specific to the reverse KL estimator. Whenever the regularizer is placed inside an expectation over trajectories, differentiating the objective produces two contributions. The first is the local derivative of the penalty at the current action, $-f'(u_i)u_i$. The second is a score-function term for all future penalties, $\sum_{j\ge i} f(u_j)$, because changing $a_i$ changes the distribution of all later prefixes. In bandit problems these two notions collapse, but in autoregressive language generation they are different. Therefore, any token-wise regularizer that is optimized as an on-policy trajectory objective induces a causal cost-to-go, and the right place to add this cost-to-go is on the regularizer side of the $\ell'$-asymmetry — i.e., \emph{after} advantage construction, as a token-level correction that is \emph{not} subjected to $\ell'$.

\subsection{Specialization to GRPO and Reverse KL}

\begin{corollary}[GRPO advantage as the score of the arcsin utility]\label{cor:grpo_arcsin_score}
Theorem~\ref{thm:f_divergence_pg} specializes to binary-reward GRPO via $\ell(p)=2\arcsin\sqrt p$, with the population GRPO advantage $A_i^{\rm GRPO}=(r(x,a)-p_\theta(x))/\sqrt{p_\theta(x)(1-p_\theta(x))}=\ell'(p_\theta(x))(r(x,a)-p_\theta(x))$ recovered as the score of $\ell$. Since $\reg_i$ is unmultiplied by $\ell'$, any reward-side modification is rescaled by $\ell'(p)$ together with $r$ while a genuine token-level regularizer is not, so the only placement consistent with both signals is to insert the regularization term \emph{after} advantage construction.
\end{corollary}
The GRPO advantage and the future-KL correction are thus objects of \emph{different scaling type} under $\ell$ and cannot be merged into a single reward without distorting one of them. Section~\ref{sec:frpo_optimality} turns this into a one-line algorithmic change.

For token-wise reverse KL, taking
\(
    f(u)=-\log u
\)
gives, up to action-independent constants that vanish under the score-function identity,
\begin{align}
    \nabla_\theta Q_{\rm rKL}(\theta)
    =
    \mathbb E_x\mathbb E_{a\sim\pi_\theta}
    \sum_{i=1}^{T}
    \left[
    A_i^{\rm GRPO}
    -
    \beta\sum_{j\ge i}
    \log\frac{\pi_\theta(a_j\mid s_j)}{\pi_{\rm ref}(a_j\mid s_j)}
    \right]
    \nabla_\theta\log\pi_\theta(a_i\mid s_i).
    \label{eq:reverse_kl_future_pg}
\end{align}
Thus each token receives the usual outcome advantage plus a causal future-KL correction. Early tokens are penalized for the regularization cost they induce downstream, not merely for their own local log-ratio.

The same theorem also covers the $k3$ generator
\(
    f(u)=-\log u+u-1.
\)
Substituting it into Eq.~\eqref{eq:kl_regularized_pg_weight} yields the coefficient
\begin{align}
    A_i^{\rm GRPO}
    -
    \beta
    \left[
    -\log u_i
    +
    \sum_{j>i}
    \left(
    -\log u_j+u_j-1
    \right)
    \right].
    \label{eq:k3_future_pg}
\end{align}
This expression differs from differentiating a local $k3$ loss term at each sampled token: the latter captures only the direct derivative of the token penalty, while Eq.~\eqref{eq:k3_future_pg} also includes the future regularization terms. The distinction matters because $k3$ was designed primarily as a low-variance nonnegative value estimator for KL. In our setting the regularizer also acts as a token-level credit signal, so the sign and centering of the cumulative term affect how the policy assigns blame or credit to earlier tokens.

\subsection{What the Loss-Side Surrogate Misses}

The classical loss-side implementation in Eq.~\eqref{eq:grpo} treats $\hat D_{i,t}$ as an additive differentiable token loss. For example,
\[
    \nabla_\theta \hat D_t^{(k3)}
    =
    (1-u_t)\nabla_\theta\log\pi_\theta(a_t\mid s_t).
\]
This is a local derivative evaluated on the sampled token and prefix. It does not differentiate the expectation over future trajectories in Eq.~\eqref{eq:general_f_objective}; consequently, it omits the $j>i$ terms in Eq.~\eqref{eq:kl_regularized_pg_weight}. This is the on-policy issue studied in this paper. It is separate from the familiar stale-rollout or off-policy issue in PPO-style training.

\begin{proposition}[Local loss-side KL omits future state-distribution terms]\label{prop:loss_side_missing_future}
Consider the on-policy objective in Eq.~\eqref{eq:general_f_objective}. A surrogate that differentiates only the sampled local penalty $f(u_i)$ at each token recovers the direct term $-f'(u_i)u_i$ but omits the score-function contribution $\sum_{j\ge i}f(u_j)$ in Eq.~\eqref{eq:kl_regularized_pg_weight}. Hence it is not the policy gradient of the trajectory-level regularized objective, except in degenerate settings where future penalties are action-independent.
\end{proposition}

For the reverse $k3$ estimator $f(u)=-\log u+u-1$, the local differentiable loss gives the coefficient $1-u_i$. The on-policy trajectory objective instead gives
\[
    \reg_i
    =
    -\log u_i
    +
    \sum_{j>i}
    \left(-\log u_j+u_j-1\right),
\]
where the second term is the missing future regularization cost. For the forward $k3$ estimator $f(u)=u\log u-u+1$, the same calculation gives
\[
    \reg_i
    =
    1-u_i
    +
    \sum_{j> i}(u_j\log u_j-u_j+1),
\]
again showing that the missing term is not an artifact of reverse KL. These expressions are the regularization analogues of reward-to-go in policy gradient: past costs can be dropped by the score-function identity, but future costs cannot.

\paragraph{Practical off-policy caveat.}
In real systems, rollouts are produced by $\pi_{\theta_{\rm old}}$ and actor updates may use mini-batches after the policy has already changed. Then even the value estimate in Proposition~\ref{prop:k3_unbiased} is no longer exactly on-policy unless importance weighting or fresh log-probabilities are used. This implementation-level mismatch can matter, but it is orthogonal to our main point: even when $\pi_{\theta_{\rm old}}=\pi_\theta$, the correct regularized policy gradient contains a future regularization return-to-go.

This distinction is useful when comparing implementations. Importance weighting can correct a mismatch between the behavior policy and the current policy in a sampled expectation, but it does not create the missing future term. Conversely, adding the future term can improve the on-policy credit assignment even when PPO epochs are set to one and rollout staleness is small. Importantly, once the future-KL correction is folded into the advantage, the whole token weight is multiplied by the PPO importance ratio $\rho_{i,t}$ in the clipped surrogate, just like the outcome advantage. This gives the regularization signal the same first-order off-policy correction as the policy-gradient term. By contrast, a separate additive loss-side KL penalty is not multiplied by $\rho_{i,t}$ in the classical surrogate, so its sampled expectation remains distribution-mismatched under stale rollouts. The two issues therefore operate on different axes: distribution correction concerns which policy generated the tokens, while future KL concerns what objective those tokens estimate.

\section{Future-KL GRPO and Optimality View}
\label{sec:frpo_optimality}

\subsection{The Decoupled FRPO Update}

Equation~\eqref{eq:reverse_kl_future_pg} suggests a lightweight modification of GRPO: compute the group-normalized outcome advantage from the original verifier reward, and then add a token-dependent future-KL correction. For response $i$ and token $t$, \method{} uses
\begin{equation}\label{eq:frpo_advantage}
    \hat A_{i,t}^{\method{}}
    =
    \operatorname{sg}\!\left(
    \hat A_i^{\rm GRPO}
    -
    \beta
    \sum_{j=t}^{T_i}
    \delta_{i,j}
    \right),
    \qquad
    \delta_{i,j}
    =
    \log\frac{\pi_{\theta}(a_{i,j}\mid s_{i,j})}
    {\pi_{\rm ref}(a_{i,j}\mid s_{i,j})}.
\end{equation}
The finite-sample implementation uses rollout log-probabilities for efficiency and then optimizes the usual clipped PPO surrogate with $\hat A_{i,t}^{\method{}}$; see Appendix~\ref{sec:app:algorithm}. Because the future-KL correction is part of the advantage, the PPO update uses $\rho_{i,t}\hat A_{i,t}^{\method{}}$ (and its clipped counterpart), so both the verifier reward signal and the regularization signal receive the same importance-ratio correction from $\pi_{\theta_{\rm old}}$ to $\pi_\theta$. This decoupled form preserves GRPO's outcome-reward normalization while restoring token-level credit assignment from the regularizer.

\paragraph{Simple change, structural reason.} Operationally, Eq.~\eqref{eq:frpo_advantage} adds about ten lines of code to a standard GRPO trainer: one reverse cumulative sum over per-token log-ratios already computed during rollout, with no extra forward passes and no learned value function. The non-trivial part is the placement, not the implementation: Corollary~\ref{cor:grpo_arcsin_score} shows that the GRPO advantage and the future-KL correction live on \emph{different sides} of the $\ell'$-asymmetry, so the only placement consistent with the on-policy gradient under non-linear $\ell$ is after advantage construction. The alternative placements fail by prediction: Proposition~\ref{prop:kl_in_reward_distortion} explains the KL-in-reward collapse observed in Section~\ref{sec:experiment}, and Proposition~\ref{prop:loss_side_missing_future} explains why loss-side KL omits the future return-to-go even on-policy.

We discuss alternative ways of inserting KL before GRPO normalization in Appendix~\ref{sec:kl_reward_normalization}. Briefly, trajectory-level KL-in-reward changes the group mean and standard deviation, while step-level normalization is fragile for small groups and variable-length responses. The decoupled update therefore keeps GRPO normalization on verifier rewards and adds future KL afterward as a token correction.



\subsection{Optimal KL-Regularized Rewards}

The same future-KL structure also appears from the optimality conditions of entropy-regularized RL. For a fixed prompt, consider
\begin{equation}\label{eq:entropy_rl}
    Q(\pi)
    =
    \mathbb E_{a\sim\pi(\cdot|x)}[r(x,a)]
    -
    \beta\,\KL[\pi(\cdot|x)\|\pi_{\rm ref}(\cdot|x)].
\end{equation}

\begin{theorem}[Optimal policy]\label{prop:prl_optimal}
The maximizer satisfies
\(
    \pi^*(a\mid x)
    \propto
    \pi_{\rm ref}(a\mid x)\exp(r(x,a)/\beta).
\)
\end{theorem}

\begin{proposition}[Constant shifted reward]\label{prop:prl_constant}
Equivalently, for each prompt $x$ there exists a constant $C(x)$ such that
\begin{equation}\label{eq:prl_constant}
    r(x,a)
    -
    \beta\log\frac{\pi^*(a\mid x)}{\pi_{\rm ref}(a\mid x)}
    =
    C(x).
\end{equation}
\end{proposition}
Because autoregressive likelihood ratios decompose over tokens, Eq.~\eqref{eq:prl_constant} implies the following process-reward form.

\begin{theorem}[Future KL as process reward]\label{prop:prl_process}
At the optimal policy, the entropy-regularized process reward after prefix $a^{(t)}$ can be written as
\begin{equation}\label{eq:prl_process}
    r_t(x,a^{(t)})
    =
    r(x,a)
    -
    \beta\sum_{j=t+1}^{T}
    \log\frac{\pi^*(a_j\mid x,a^{(j-1)})}
    {\pi_{\rm ref}(a_j\mid x,a^{(j-1)})}.
\end{equation}
\end{theorem}

Proofs are deferred to Appendix~\ref{sec:app:proofs:prl}. This optimality view is not needed to derive Theorem~\ref{thm:f_divergence_pg}, but it gives the same interpretation: KL regularization is a process-level signal, and the relevant token weight is a future cost-to-go rather than an isolated per-token penalty.

The optimality view also explains why the future term is naturally causal. For two adjacent prefixes, Eq.~\eqref{eq:prl_process} implies
\[
    r_t(x,a^{(t)})-r_{t-1}(x,a^{(t-1)})
    =
    \beta
    \log\frac{\pi^*(a_t\mid x,a^{(t-1)})}
    {\pi_{\rm ref}(a_t\mid x,a^{(t-1)})}.
\]
Thus the change in process reward between consecutive steps is exactly a token-level log-ratio at the selected action. The future-KL correction in \method{} can be viewed as a practical on-policy approximation to this structure, with $\pi_{\theta_{\rm old}}$ used for rollout log-probabilities and GRPO's group-normalized advantage used in place of a learned process value. This gives a direct bridge between the policy-gradient derivation and the process-level interpretation of KL-regularized optimality.

\section{Experiments and Results}
\label{sec:experiment}
\vspace{-0.5em}

In this section, we briefly summarize the experiments with their results. For details about hyperparameters, other configurations, and more experiment results, please refer to Appendix~\ref{sec:app:experiment}.

\noindent\textbf{Basic Configuration} We select Qwen3-1.7B-Base and Qwen3-4B-Base \citep{qwen3} for dense models, and Qwen3-30B-A3B-Base for MoE models as base models. For the training dataset, we use a filtered version of DAPO-Math-17k \citep{yu2025dapo}, which contains about 12,000 problems. For the evaluation benchmarks, we choose MATH500 \citep{hendrycks2021measuring}, AIME24, AIME25, AMC23, OlympiadBench~\citep{he2024olympiadbench}, and MinervaMath~\citep{lewkowycz2022solving}. For the metric pass@16, we perform a bootstrap by sampling with replacement 1,000 times and report the mean.

\subsection{Main Experiments}
\begin{figure}[h]
    \centering
        \includegraphics[width=\linewidth]{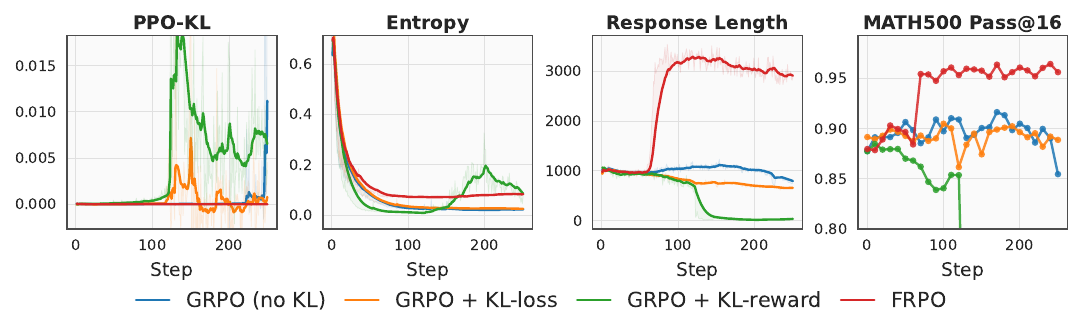}
        \vspace{-1.0em}
    \caption{The training dynamics of Qwen-30B-A3B-Base with different KL integrations.}
    \label{fig:main_30ba3b_reversekl}
\end{figure}

Figure~\ref{fig:main_30ba3b_reversekl} demonstrates that incorporating the future-KL correction leads to higher Pass@n performance in the Qwen3-30B-A3B-Base setting, outperforming the strongest baseline in this comparison by nearly 5\% absolute gain. This is accompanied by higher final entropy and lower PPO-KL, which measures policy drift from the reference policy for \method{}.
The finetuned model remains close to the reference policy as measured by PPO-KL divergence, suggesting that the base model already contains useful long-\gls{cot} reasoning patterns and that future-KL credit assignment helps elicit them without excessive drift. Evaluations on the AIME24 and AIME25 benchmarks also show that \method{} achieves better pass@n performance than using KL in loss directly for the large model. The Qwen3-1.7B-Base result is weaker, and we hypothesize that this model may be closer to its reasoning capacity under the current training budget and hyperparameters.

\begin{table}[h]
    \centering
    \caption{Full evaluation results of pass@16 (\%) for Qwen3-1.7-Base, Qwen3-4B-Base, and Qwen3-30B-A3B-Base on different benchmarks. Each entry reports the mean with the standard deviation in parentheses.}
    \label{tab:full_results_pass16}
    \resizebox{\textwidth}{!}{
    \begin{tabular}{lcccccc}
    \toprule
        \textbf{Model} & \textbf{MATH500} & \textbf{AIME24} & \textbf{AIME25} & \textbf{AMC23} & \textbf{OlympiadBench} & \textbf{MinervaMath} \\\midrule
        Qwen3-1.7B-Base & 69.65 (2.48) & 12.94 (2.87) & 7.15 (2.45) & 52.76 (4.01) & 38.80 (0.75) & 36.02 (1.16) \\
        \quad + GRPO (no KL) & 78.62 (2.01) & 25.82 (2.92) & 16.76 (2.72) & 69.73 (2.97) & 57.23 (0.58) & \textbf{48.60} (0.83) \\
        \quad + GRPO KL-loss & 77.07 (1.68) & \textbf{27.15} (2.65) & 18.33 (3.26) & \textbf{81.56} (3.23) & 54.90 (0.56) & 46.36 (0.90) \\
        \quad + GRPO KL-reward & collapse & / & / & / & / & / \\
        \quad + \method{} & \textbf{85.30} (3.00) & {27.00} (2.53) & \textbf{20.31} (3.73) & 74.47 (3.00) & \textbf{59.13} (0.65) & {47.15} (1.03) \\\midrule
        Qwen3-4B-Base & 70.99 (1.08) & 21.08 (2.37) & 22.26 (3.17) & 71.39 (3.27) & 52.27 (0.68) & 46.98 (0.90) \\
        \quad + GRPO (no KL) & 86.15 (0.85) & 29.43 (2.87) & 34.16 (2.48) & 85.73 (2.67) & 64.28 (0.51) & 55.07 (0.79) \\
        \quad + GRPO KL-loss & 91.11 (1.36) & \textbf{46.53} (2.83) & \textbf{50.89} (3.65) & 90.10 (2.11) & \textbf{72.50} (0.48) & \textbf{58.64} (0.87) \\
        \quad + GRPO KL-reward & collapse & / & / & / & / & / \\
        \quad + \method{} & \textbf{92.93} (1.70) & 41.09 (3.43) & 42.17 (4.03) & \textbf{92.41} (2.16) & 72.23 (0.51) & 58.56 (0.76) \\\midrule
        Qwen3-30B-A3B-Base & 87.77 (2.08) & 35.31 (3.38) & 21.04 (2.39) & 76.67 (2.44) & 61.36 (0.76) & 51.68 (0.95) \\
        \quad + GRPO (no KL) & 91.63 (0.87) & 53.06 (2.99) & 27.45 (2.30) & 91.44 (1.31) & 66.33 (0.49) & 57.46 (0.68) \\
        \quad + GRPO KL-loss & 90.23 (1.17) & 51.25 (4.52) & 22.04 (3.05) & 89.06 (2.57) & 62.86 (0.49) & 55.51 (0.70) \\
        \quad + GRPO KL-reward & collapse & / & / & / & / & / \\
        \quad + \method{} & \textbf{96.07} (1.54) & \textbf{63.88} (3.86) & \textbf{45.23} (3.41) & \textbf{94.25} (1.35) & \textbf{77.71} (0.47) & \textbf{58.65} (0.71) \\
        \bottomrule
    \end{tabular}}
\end{table}

\subsection{Ablation Studies}
\subsubsection{KL Direction and Estimator Ablations}
\label{sec:app:kl_direction_estimator}

\begin{figure}[h]
    \centering
    \includegraphics[width=\linewidth]{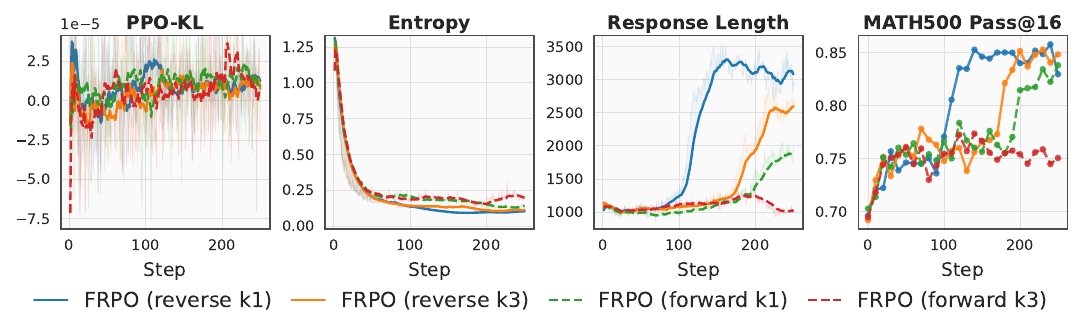}
    \caption{Reverse KL versus forward KL, with $k1$ and $k3$ as the estimators on Qwen3-1.7B-Base.}
    \label{fig:reverse_forward_k1_k3}
\end{figure}

\noindent\textbf{Forward KL versus Reverse KL.}
Forward KL $\KL(\pi_{\rm ref}\|\pi_\theta)$ is mode-covering: it penalizes the current policy for assigning too little probability to regions supported by the reference policy. Reverse KL $\KL(\pi_\theta\|\pi_{\rm ref})$ is more mode-seeking: it discourages generations that are unlikely under the reference. Figure~\ref{fig:reverse_forward_k1_k3} shows that reverse KL works better in our setting, likely because forward KL keeps entropy higher than is useful for math reasoning.

\noindent\textbf{Different KL Estimators.}
Many \gls{rl} frameworks favor $k3$ over $k1$ because both are value-unbiased for reverse KL under on-policy sampling, while $k3$ is always nonnegative. In \method{}, however, the future-KL term is used as a signed credit-assignment signal rather than only as a scalar divergence monitor. The nonnegativity of $k3$ makes its cumulative future term one-sided: it always pushes the token weight in the penalty direction, even when a centered signal would be needed to distinguish relatively helpful and harmful deviations within a group. Figure~\ref{fig:reverse_forward_k1_k3} shows a gap between $k1$ and $k3$ estimators, especially on forward KL, so we use $k1$ as the default estimator in \method{}.

\subsubsection{Subtracting a Baseline for $k3$ Estimators}
Due to the nonnegativity of $k3$ estimators, the future KL term will always be a penalty, therefore pushing the policy towards the same direction. To avoid such a phenomenon, we implement a naive baseline, which takes the average KL divergence at the token level among all tokens in the same batch, and then scaled by the remaining length of a particular response from the current token $i$, i.e., 
\begin{align*}
    b_i=(T-i)\cdot \overline{\KL}.
\end{align*}
In Table~\ref{tab:k3_with_baselines}, we summarize the results of incorporating baseline subtraction into the $k3$ estimators for both forward and reverse KL. We observe that baseline subtraction alone does not lead to performance improvements. We hypothesize that this is because the estimated baselines are relatively coarse and do not accurately approximate the true baselines. More sophisticated baseline estimation methods may be beneficial, and we leave their investigation to future work.

\begin{table}[h]
    \centering
    \caption{Ablation studies with subtracting a baseline from the nonnegative $k3$ KL estimator. Experiments are done on Qwen3-1.7B-Base. Performances are measured by Pass@16 (\%).}
    \label{tab:k3_with_baselines}
    \resizebox{\textwidth}{!}{
    \begin{tabular}{lcccccc}
    \toprule
        \textbf{Method} & \textbf{MATH500} & \textbf{AIME24} & \textbf{AIME25} & \textbf{AMC23} & \textbf{OlympiadBench} & \textbf{MinervaMath} \\\midrule
        Forward $k3$ with baseline & 75.21 (2.37) & 20.68 (2.39) & 10.45 (2.52) & \textbf{72.66} (2.95) & 52.50 (0.60) & \textbf{46.86} (0.90) \\
        Forward $k3$ without baseline & \textbf{75.62} (2.55) & \textbf{22.32} (3.22) & \textbf{18.94} (3.31) & 70.55 (2.84) & \textbf{53.05} (0.66) & 46.37 (0.98) \\
        Reverse $k3$ with baseline & 68.27 (0.75) & 20.63 (2.55) & \textbf{19.89} (2.89) & 76.36 (3.23) & 54.14 (0.60) & 47.82 (1.01)\\
        Reverse $k3$ without baseline & \textbf{84.82} (3.43) & \textbf{32.23} (4.29) & 11.76 (1.91) & \textbf{79.16} (3.83) & \textbf{58.57} (0.62) & \textbf{48.02} (1.02) \\
        \bottomrule
    \end{tabular}}
\end{table}

\subsubsection{Ablation for Different Advantage Estimators}
\begin{figure}[h]
    \centering
    \includegraphics[width=1\linewidth]{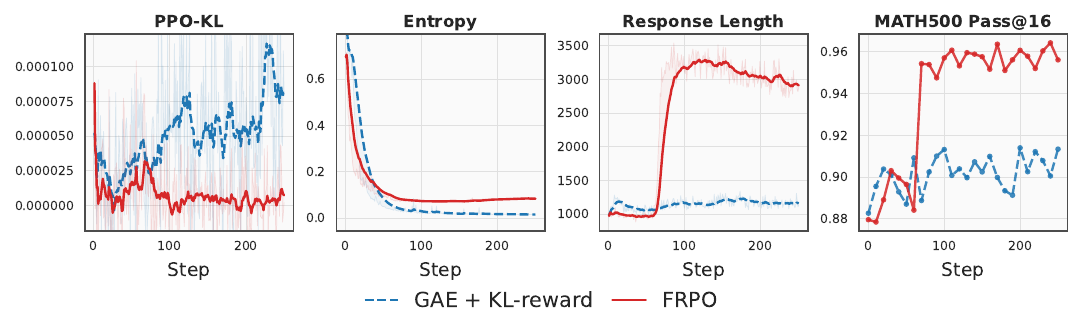}
    \caption{Comparison between GAE and GRPO advantage estimation, both with future KL.}
    \label{fig:30ba3b_gae_vs_frpo}
    \vspace{-1em}
\end{figure}

In existing \gls{rl} frameworks such as verl \citep{sheng2024hybridflow} and AReaL \citep{fu2025areal}, KL-in-reward can be combined with the GAE advantage estimator. This folds the regularizer into token rewards before advantage estimation. In contrast, \method{} preserves the \gls{grpo}-style group-normalized outcome advantage and adds future KL afterward. Figure~\ref{fig:30ba3b_gae_vs_frpo}, based on Qwen3-30B-A3B-Base, suggests that retaining GRPO's outcome-level normalization is beneficial compared with integrating KL into rewards before estimating advantages.

Other ablation studies and full evaluation results could be found in Appendix~\ref{sec:app:experiment}.

\section{Conclusion and Discussion}
\vspace{-0.5em}

We revisited KL regularization in GRPO from the on-policy objective rather than only from the implementation of a local loss penalty. The resulting policy-gradient formula applies to general token-wise $f$-divergence regularizers and shows that autoregressive regularization induces a future return-to-go term. For reverse KL, this gives the future-KL correction used by \method{}, which keeps GRPO's group-normalized outcome advantage separate from token-level regularization credit. The same structure is supported by the optimality view of KL-regularized trajectory rewards. In mathematical-reasoning experiments, this formulation improves pass@n while limiting policy drift and entropy collapse.


\bibliographystyle{plainnat}
\bibliography{ref}

@article{shao2024deepseekmath,
  title={Deepseekmath: Pushing the limits of mathematical reasoning in open language models},
  author={Shao, Zhihong and Wang, Peiyi and Zhu, Qihao and Xu, Runxin and Song, Junxiao and Bi, Xiao and Zhang, Haowei and Zhang, Mingchuan and Li, YK and Wu, Yang and others},
  journal={arXiv preprint arXiv:2402.03300},
  year={2024}
}

@article{guo2025deepseek,
  title={Deepseek-r1: Incentivizing reasoning capability in llms via reinforcement learning},
  author={Guo, Daya and Yang, Dejian and Zhang, Haowei and Song, Junxiao and Wang, Peiyi and Zhu, Qihao and Xu, Runxin and Zhang, Ruoyu and Ma, Shirong and Bi, Xiao and others},
  journal={arXiv preprint arXiv:2501.12948},
  year={2025}
}

@article{ouyang2022training,
  title={Training language models to follow instructions with human feedback},
  author={Ouyang, Long and Wu, Jeffrey and Jiang, Xu and Almeida, Diogo and Wainwright, Carroll and Mishkin, Pamela and Zhang, Chong and Agarwal, Sandhini and Slama, Katarina and Ray, Alex and others},
  journal={Advances in neural information processing systems},
  volume={35},
  pages={27730--27744},
  year={2022}
}

@misc{schulman2020klapprox,
  author       = {John Schulman},
  title        = {Approximating KL Divergence},
  year         = {2020},
  howpublished = {\url{http://joschu.net/blog/kl-approx.html}},
  note         = {Blog post}
}

@article{yao2025optimizing,
  title={Optimizing Chain-of-Thought Reasoners via Gradient Variance Minimization in Rejection Sampling and RL},
  author={Yao, Jiarui and Hao, Yifan and Zhang, Hanning and Dong, Hanze and Xiong, Wei and Jiang, Nan and Zhang, Tong},
  journal={arXiv preprint arXiv:2505.02391},
  year={2025}
}

@article{xiong2025minimalist,
  title={A minimalist approach to llm reasoning: from rejection sampling to reinforce},
  author={Xiong, Wei and Yao, Jiarui and Xu, Yuhui and Pang, Bo and Wang, Lei and Sahoo, Doyen and Li, Junnan and Jiang, Nan and Zhang, Tong and Xiong, Caiming and others},
  journal={arXiv preprint arXiv:2504.11343},
  year={2025}
}

@article{zheng2025group,
  title={Group sequence policy optimization},
  author={Zheng, Chujie and Liu, Shixuan and Li, Mingze and Chen, Xiong-Hui and Yu, Bowen and Gao, Chang and Dang, Kai and Liu, Yuqiong and Men, Rui and Yang, An and others},
  journal={arXiv preprint arXiv:2507.18071},
  year={2025}
}

@article{liu2026gdpo,
  title={Gdpo: Group reward-decoupled normalization policy optimization for multi-reward rl optimization},
  author={Liu, Shih-Yang and Dong, Xin and Lu, Ximing and Diao, Shizhe and Belcak, Peter and Liu, Mingjie and Chen, Min-Hung and Yin, Hongxu and Wang, Yu-Chiang Frank and Cheng, Kwang-Ting and others},
  journal={arXiv preprint arXiv:2601.05242},
  year={2026}
}

@inproceedings{kwon2023efficient,
  title={Efficient Memory Management for Large Language Model Serving with PagedAttention},
  author={Woosuk Kwon and Zhuohan Li and Siyuan Zhuang and Ying Sheng and Lianmin Zheng and Cody Hao Yu and Joseph E. Gonzalez and Hao Zhang and Ion Stoica},
  booktitle={Proceedings of the ACM SIGOPS 29th Symposium on Operating Systems Principles},
  year={2023}
}

@article{yang2024qwen2,
  title={Qwen2. 5-math technical report: Toward mathematical expert model via self-improvement},
  author={Yang, An and Zhang, Beichen and Hui, Binyuan and Gao, Bofei and Yu, Bowen and Li, Chengpeng and Liu, Dayiheng and Tu, Jianhong and Zhou, Jingren and Lin, Junyang and others},
  journal={arXiv preprint arXiv:2409.12122},
  year={2024}
}

@article{qwen3,
    title={Qwen3 Technical Report}, 
    author={An Yang and Anfeng Li and Baosong Yang and Beichen Zhang and Binyuan Hui and Bo Zheng and Bowen Yu and Chang Gao and Chengen Huang and Chenxu Lv and Chujie Zheng and Dayiheng Liu and Fan Zhou and Fei Huang and Feng Hu and Hao Ge and Haoran Wei and Huan Lin and Jialong Tang and Jian Yang and Jianhong Tu and Jianwei Zhang and Jianxin Yang and Jiaxi Yang and Jing Zhou and Jingren Zhou and Junyang Lin and Kai Dang and Keqin Bao and Kexin Yang and Le Yu and Lianghao Deng and Mei Li and Mingfeng Xue and Mingze Li and Pei Zhang and Peng Wang and Qin Zhu and Rui Men and Ruize Gao and Shixuan Liu and Shuang Luo and Tianhao Li and Tianyi Tang and Wenbiao Yin and Xingzhang Ren and Xinyu Wang and Xinyu Zhang and Xuancheng Ren and Yang Fan and Yang Su and Yichang Zhang and Yinger Zhang and Yu Wan and Yuqiong Liu and Zekun Wang and Zeyu Cui and Zhenru Zhang and Zhipeng Zhou and Zihan Qiu},
    journal = {arXiv preprint arXiv:2505.09388},
    year={2025}
}

@inproceedings{vieillard2020momentum,
  title={Momentum in reinforcement learning},
  author={Vieillard, Nino and Scherrer, Bruno and Pietquin, Olivier and Geist, Matthieu},
  booktitle={International Conference on Artificial Intelligence and Statistics},
  pages={2529--2538},
  year={2020},
  organization={PMLR}
}

@article{della2022entropy,
  title={Entropy regularized reinforcement learning with cascading networks},
  author={Della Vecchia, Riccardo and Shilova, Alena and Preux, Philippe and Akrour, Riad},
  journal={arXiv preprint arXiv:2210.08503},
  year={2022}
}

@article{eysenbach2018diversity,
  title={Diversity is all you need: Learning skills without a reward function},
  author={Eysenbach, Benjamin and Gupta, Abhishek and Ibarz, Julian and Levine, Sergey},
  journal={arXiv preprint arXiv:1802.06070},
  year={2018}
}

@article{bai2022training,
  title={Training a helpful and harmless assistant with reinforcement learning from human feedback},
  author={Bai, Yuntao and Jones, Andy and Ndousse, Kamal and Askell, Amanda and Chen, Anna and DasSarma, Nova and Drain, Dawn and Fort, Stanislav and Ganguli, Deep and Henighan, Tom and others},
  journal={arXiv preprint arXiv:2204.05862},
  year={2022}
}

@inproceedings{ahn2024large,
  title={Large language models for mathematical reasoning: Progresses and challenges},
  author={Ahn, Janice and Verma, Rishu and Lou, Renze and Liu, Di and Zhang, Rui and Yin, Wenpeng},
  booktitle={Proceedings of the 18th Conference of the European Chapter of the Association for Computational Linguistics: Student Research Workshop},
  pages={225--237},
  year={2024}
}

@article{hendrycks2021measuring,
  title={Measuring mathematical problem solving with the math dataset},
  author={Hendrycks, Dan and Burns, Collin and Kadavath, Saurav and Arora, Akul and Basart, Steven and Tang, Eric and Song, Dawn and Steinhardt, Jacob},
  journal={arXiv preprint arXiv:2103.03874},
  year={2021}
}

@inproceedings{lightman2023let,
  title={Let's verify step by step},
  author={Lightman, Hunter and Kosaraju, Vineet and Burda, Yuri and Edwards, Harrison and Baker, Bowen and Lee, Teddy and Leike, Jan and Schulman, John and Sutskever, Ilya and Cobbe, Karl},
  booktitle={The twelfth international conference on learning representations},
  year={2023}
}

@article{liu2025understanding,
  title={Understanding r1-zero-like training: A critical perspective},
  author={Liu, Zichen and Chen, Changyu and Li, Wenjun and Qi, Penghui and Pang, Tianyu and Du, Chao and Lee, Wee Sun and Lin, Min},
  journal={arXiv preprint arXiv:2503.20783},
  year={2025}
}

@inproceedings{he2024olympiadbench,
  title={Olympiadbench: A challenging benchmark for promoting agi with olympiad-level bilingual multimodal scientific problems},
  author={He, Chaoqun and Luo, Renjie and Bai, Yuzhuo and Hu, Shengding and Thai, Zhen and Shen, Junhao and Hu, Jinyi and Han, Xu and Huang, Yujie and Zhang, Yuxiang and others},
  booktitle={Proceedings of the 62nd Annual Meeting of the Association for Computational Linguistics (Volume 1: Long Papers)},
  pages={3828--3850},
  year={2024}
}

@article{lewkowycz2022solving,
  title={Solving quantitative reasoning problems with language models},
  author={Lewkowycz, Aitor and Andreassen, Anders and Dohan, David and Dyer, Ethan and Michalewski, Henryk and Ramasesh, Vinay and Slone, Ambrose and Anil, Cem and Schlag, Imanol and Gutman-Solo, Theo and others},
  journal={Advances in neural information processing systems},
  volume={35},
  pages={3843--3857},
  year={2022}
}

@article{liu2019policy,
  title={Policy optimization reinforcement learning with entropy regularization},
  author={Liu, Jingbin and Gu, Xinyang and Liu, Shuai},
  journal={arXiv preprint arXiv:1912.01557},
  year={2019}
}

@misc{numina_math_datasets,
  author = {Jia LI and Edward Beeching and Lewis Tunstall and Ben Lipkin and Roman Soletskyi and Shengyi Costa Huang and Kashif Rasul and Longhui Yu and Albert Jiang and Ziju Shen and Zihan Qin and Bin Dong and Li Zhou and Yann Fleureau and Guillaume Lample and Stanislas Polu},
  title = {NuminaMath},
  year = {2024},
  publisher = {Numina},
  journal = {Hugging Face repository},
  howpublished = {\url{[https://huggingface.co/AI-MO/NuminaMath-CoT](https://github.com/project-numina/aimo-progress-prize/blob/main/report/numina_dataset.pdf)}}
}

@article{sheng2024hybridflow,
  title   = {HybridFlow: A Flexible and Efficient RLHF Framework},
  author  = {Guangming Sheng and Chi Zhang and Zilingfeng Ye and Xibin Wu and Wang Zhang and Ru Zhang and Yanghua Peng and Haibin Lin and Chuan Wu},
  year    = {2024},
  journal = {arXiv preprint arXiv: 2409.19256}
}

@article{wen2024entropy,
  title={Entropy-regularized token-level policy optimization for language agent reinforcement},
  author={Wen, Muning and Liao, Junwei and Deng, Cheng and Wang, Jun and Zhang, Weinan and Wen, Ying},
  journal={arXiv preprint arXiv:2402.06700},
  year={2024}
}

@misc{yao2025offpolicy,
  title = {Your Efficient RL Framework Secretly Brings You Off-Policy RL Training},
  url = {https://fengyao.notion.site/off-policy-rl},
  author = {Yao, Feng and Liu, Liyuan and Zhang, Dinghuai and Dong, Chengyu and Shang, Jingbo and Gao, Jianfeng},
  journal = {Feng Yao's Notion},
  year = {2025},
  month = aug,
}

@article{yuan2024free,
  title={Free process rewards without process labels},
  author={Yuan, Lifan and Li, Wendi and Chen, Huayu and Cui, Ganqu and Ding, Ning and Zhang, Kaiyan and Zhou, Bowen and Liu, Zhiyuan and Peng, Hao},
  journal={arXiv preprint arXiv:2412.01981},
  year={2024}
}

@article{hu2025reinforce++,
  title={Reinforce++: Stabilizing critic-free policy optimization with global advantage normalization},
  author={Hu, Jian and Liu, Jason Klein and Xu, Haotian and Shen, Wei},
  journal={arXiv preprint arXiv:2501.03262},
  year={2025}
}

@article{kumar2024training,
  title={Training language models to self-correct via reinforcement learning},
  author={Kumar, Aviral and Zhuang, Vincent and Agarwal, Rishabh and Su, Yi and Co-Reyes, John D and Singh, Avi and Baumli, Kate and Iqbal, Shariq and Bishop, Colton and Roelofs, Rebecca and others},
  journal={arXiv preprint arXiv:2409.12917},
  year={2024}
}

@article{cui2025process,
  title={Process reinforcement through implicit rewards},
  author={Cui, Ganqu and Yuan, Lifan and Wang, Zefan and Wang, Hanbin and Li, Wendi and He, Bingxiang and Fan, Yuchen and Yu, Tianyu and Xu, Qixin and Chen, Weize and others},
  journal={arXiv preprint arXiv:2502.01456},
  year={2025}
}

@article{jin2025search,
  title={Search-r1: Training llms to reason and leverage search engines with reinforcement learning},
  author={Jin, Bowen and Zeng, Hansi and Yue, Zhenrui and Yoon, Jinsung and Arik, Sercan and Wang, Dong and Zamani, Hamed and Han, Jiawei},
  journal={arXiv preprint arXiv:2503.09516},
  year={2025}
}

@article{wei2022emergent,
  title={Emergent abilities of large language models},
  author={Wei, Jason and Tay, Yi and Bommasani, Rishi and Raffel, Colin and Zoph, Barret and Borgeaud, Sebastian and Yogatama, Dani and Bosma, Maarten and Zhou, Denny and Metzler, Donald and others},
  journal={arXiv preprint arXiv:2206.07682},
  year={2022}
}

@article{qi2025defeating,
  title={Defeating the training-inference mismatch via fp16},
  author={Qi, Penghui and Liu, Zichen and Zhou, Xiangxin and Pang, Tianyu and Du, Chao and Lee, Wee Sun and Lin, Min},
  journal={arXiv preprint arXiv:2510.26788},
  year={2025}
}

@article{yu2025dapo,
  title={Dapo: An open-source llm reinforcement learning system at scale},
  author={Yu, Qiying and Zhang, Zheng and Zhu, Ruofei and Yuan, Yufeng and Zuo, Xiaochen and Yue, Yu and Dai, Weinan and Fan, Tiantian and Liu, Gaohong and Liu, Lingjun and others},
  journal={arXiv preprint arXiv:2503.14476},
  year={2025}
}

@misc{fu2025areal,
      title={AReaL: A Large-Scale Asynchronous Reinforcement Learning System for Language Reasoning},
      author={Wei Fu and Jiaxuan Gao and Xujie Shen and Chen Zhu and Zhiyu Mei and Chuyi He and Shusheng Xu and Guo Wei and Jun Mei and Jiashu Wang and Tongkai Yang and Binhang Yuan and Yi Wu},
      year={2025},
      eprint={2505.24298},
      archivePrefix={arXiv},
      primaryClass={cs.LG},
      url={https://arxiv.org/abs/2505.24298},
}

@article{wei2022chain,
  title={Chain-of-thought prompting elicits reasoning in large language models},
  author={Wei, Jason and Wang, Xuezhi and Schuurmans, Dale and Bosma, Maarten and Xia, Fei and Chi, Ed and Le, Quoc V and Zhou, Denny and others},
  journal={Advances in neural information processing systems},
  volume={35},
  pages={24824--24837},
  year={2022}
}

@article{zhang2025design,
  title={On the design of kl-regularized policy gradient algorithms for llm reasoning},
  author={Zhang, Yifan and Liu, Yifeng and Yuan, Huizhuo and Yuan, Yang and Gu, Quanquan and Yao, Andrew Chi-Chih},
  journal={arXiv preprint arXiv:2505.17508},
  year={2025}
}

@article{zhao2024sharp,
  title={Sharp analysis for kl-regularized contextual bandits and rlhf},
  author={Zhao, Heyang and Ye, Chenlu and Gu, Quanquan and Zhang, Tong},
  journal={arXiv preprint arXiv:2411.04625},
  year={2024}
}

@article{schulman2017proximal,
  title={Proximal policy optimization algorithms},
  author={Schulman, John and Wolski, Filip and Dhariwal, Prafulla and Radford, Alec and Klimov, Oleg},
  journal={arXiv preprint arXiv:1707.06347},
  year={2017}
}

@inproceedings{schulman2015trust,
  title={Trust region policy optimization},
  author={Schulman, John and Levine, Sergey and Abbeel, Pieter and Jordan, Michael and Moritz, Philipp},
  booktitle={International conference on machine learning},
  pages={1889--1897},
  year={2015},
  organization={PMLR}
}

@inproceedings{tiapkin2024generative,
  title={Generative flow networks as entropy-regularized rl},
  author={Tiapkin, Daniil and Morozov, Nikita and Naumov, Alexey and Vetrov, Dmitry P},
  booktitle={International Conference on Artificial Intelligence and Statistics},
  pages={4213--4221},
  year={2024},
  organization={PMLR}
}

@article{kozuno2022kl,
  title={Kl-entropy-regularized rl with a generative model is minimax optimal},
  author={Kozuno, Tadashi and Yang, Wenhao and Vieillard, Nino and Kitamura, Toshinori and Tang, Yunhao and Mei, Jincheng and M{\'e}nard, Pierre and Azar, Mohammad Gheshlaghi and Valko, Michal and Munos, R{\'e}mi and others},
  journal={arXiv preprint arXiv:2205.14211},
  year={2022}
}

@article{luo2026ratio,
  title={Ratio-Variance Regularized Policy Optimization for Efficient LLM Fine-tuning},
  author={Luo, Yu and Han, Shuo and Hu, Yihan and Li, Dong and Hao, Jianye},
  journal={arXiv preprint arXiv:2601.03320},
  year={2026}
}

@article{jaech2024openai,
  title={Openai o1 system card},
  author={Jaech, Aaron and Kalai, Adam and Lerer, Adam and Richardson, Adam and El-Kishky, Ahmed and Low, Aiden and Helyar, Alec and Madry, Aleksander and Beutel, Alex and Carney, Alex and others},
  journal={arXiv preprint arXiv:2412.16720},
  year={2024}
}

@article{rafailov2023direct,
  title={Direct preference optimization: Your language model is secretly a reward model},
  author={Rafailov, Rafael and Sharma, Archit and Mitchell, Eric and Manning, Christopher D and Ermon, Stefano and Finn, Chelsea},
  journal={Advances in neural information processing systems},
  volume={36},
  pages={53728--53741},
  year={2023}
}

@article{huang2024correcting,
  title={Correcting the mythos of kl-regularization: Direct alignment without overoptimization via chi-squared preference optimization},
  author={Huang, Audrey and Zhan, Wenhao and Xie, Tengyang and Lee, Jason D and Sun, Wen and Krishnamurthy, Akshay and Foster, Dylan J},
  journal={arXiv preprint arXiv:2407.13399},
  year={2024}
}

@article{xiong2023iterative,
  title={Iterative preference learning from human feedback: Bridging theory and practice for rlhf under kl-constraint},
  author={Xiong, Wei and Dong, Hanze and Ye, Chenlu and Wang, Ziqi and Zhong, Han and Ji, Heng and Jiang, Nan and Zhang, Tong},
  journal={arXiv preprint arXiv:2312.11456},
  year={2023}
}

@article{vieillard2020leverage,
  title={Leverage the average: an analysis of kl regularization in reinforcement learning},
  author={Vieillard, Nino and Kozuno, Tadashi and Scherrer, Bruno and Pietquin, Olivier and Munos, R{\'e}mi and Geist, Matthieu},
  journal={Advances in Neural Information Processing Systems},
  volume={33},
  pages={12163--12174},
  year={2020}
}

@article{yang2026your,
  title={Your Group-Relative Advantage Is Biased},
  author={Yang, Fengkai and Chen, Zherui and Wang, Xiaohan and Lu, Xiaodong and Chai, Jiajun and Yin, Guojun and Lin, Wei and Ma, Shuai and Zhuang, Fuzhen and Wang, Deqing and others},
  journal={arXiv preprint arXiv:2601.08521},
  year={2026}
}

@article{qi2026rethinking,
  title={Rethinking the Trust Region in LLM Reinforcement Learning},
  author={Qi, Penghui and Zhou, Xiangxin and Liu, Zichen and Pang, Tianyu and Du, Chao and Lin, Min and Lee, Wee Sun},
  journal={arXiv preprint arXiv:2602.04879},
  year={2026}
}

@article{xiong2025reinforce,
  title={Reinforce-Ada: An Adaptive Sampling Framework under Non-linear RL Objectives},
  author={Xiong, Wei and Ye, Chenlu and Liao, Baohao and Dong, Hanze and Xu, Xinxing and Monz, Christof and Bian, Jiang and Jiang, Nan and Zhang, Tong},
  journal={arXiv preprint arXiv:2510.04996},
  year={2025}
}

@inproceedings{zhang2025learning,
  title={Learning like humans: Advancing llm reasoning capabilities via adaptive difficulty curriculum learning and expert-guided self-reformulation},
  author={Zhang, Enci and Yan, Xingang and Lin, Wei and Zhang, Tianxiang and Qianchun, Lu},
  booktitle={Proceedings of the 2025 Conference on Empirical Methods in Natural Language Processing},
  pages={6630--6644},
  year={2025}
}


\clearpage
\appendix

\section{Related Work}

\paragraph{\gls{rl} for \gls{llm} Policy Optimization} In the classical \gls{rl} domain, TRPO \citep{schulman2015trust} and PPO \citep{schulman2017proximal} are generally utilized to optimize models. As \glspl{llm} become popular, \gls{rl} algorithms are applied to further improve their abilities \citep{rafailov2023direct,jaech2024openai,ouyang2022training,bai2022training}. Since the release of DeepSeek-R1 \citep{guo2025deepseek}, critic-free policy optimization \citep{liu2025understanding,hu2025reinforce++} has become more and more popular in optimizing \gls{llm}, as it could be formulated into a bandit problem in a vast number of scenarios, leading to simple adaptations to many training settings. Based on the initially proposed \gls{grpo} \citep{shao2024deepseekmath,guo2025deepseek} algorithm, many follow-up variants appear to resolve the bias \citep{liu2025understanding,yang2026your}, mitigate the variance \citep{qi2026rethinking,luo2026ratio}, improve the sample efficiency \citep{yao2025optimizing,xiong2025reinforce}, and derive simpler and more effective algorithms \citep{xiong2025minimalist,yu2025dapo,zhang2025learning,liu2026gdpo,zheng2025group}.

\paragraph{Entropy-Regularized \gls{rl}} From the theoretical perspective, \citep{zhao2024sharp} establishes a sharp convergence rate compared to normal \gls{rl} without regularization. Other works \citep{huang2024correcting,xiong2023iterative,vieillard2020leverage,liu2019policy,wen2024entropy,eysenbach2018diversity,della2022entropy,vieillard2020momentum} also demonstrate the effectiveness of regularization in \gls{rl} optimization both from mathematical analysis and empirical results. Besides, entropy-regularized \gls{rl} is also an important technique in training models other than \gls{llm}, e.g., \citet{tiapkin2024generative}, \citet{kozuno2022kl}, as it limits the deviation from the base model when it is strong enough and maintains the generalizability on domains not \gls{rl} finetuned on.

\paragraph{\gls{llm} for Reasoning \gls{rl}} Due to the strong capabilities of pretrained \glspl{llm}, they have been used to tackle challenging scenarios requiring complex reasoning \citep{wei2022emergent,jin2025search,kumar2024training}. Among various reasoning domains, mathematical reasoning is particularly important due to its ease of verification and its foundational role for other tasks \citep{wei2022chain,shao2024deepseekmath,yang2024qwen2,ahn2024large,lightman2023let,ahn2024large,yuan2024free,cui2025process}. The base models, evaluation metrics, and other supporting techniques are relatively mature; therefore, we also select math reasoning as the empirical task to verify the effectiveness of our proposed method. Commonly used benchmarks \citep{hendrycks2021measuring,he2024olympiadbench,lewkowycz2022solving} and open-sourced datasets \citep{yu2025dapo,numina_math_datasets} are also available for conducting large-scale experiments. 
\section{KL Estimators}
\label{sec:app:kl_estimators}

\subsection{Reverse KL Estimator Variants and Their Properties}

Following \cite{schulman2020klapprox}, we analyze four estimators. Since the original \gls{grpo} objective and popular RL frameworks like verl use reverse KL as the default option, we first summarize the expectations and gradients for reverse KL-based estimators, i.e., $\KL(\pi_\theta\|\pi_{\rm ref})$.

\begin{definition}[KL Estimators]\label{def:estimators}
\begin{alignat*}{2}
    &\hat{D}^{(k1)} &&= \delta = \log \alpha, \\
    &\hat{D}^{(k2)} &&= \tfrac{1}{2}\delta^2 = \tfrac{1}{2}(\log \alpha)^2, \\
    &\hat{D}^{(k3)} &&= \alpha^{-1} + \delta - 1 = \alpha^{-1} + \log \alpha - 1, \\
    &\hat{D}^{(\text{full})} &&= \sum_{a \in \mathcal{V}} \pi_\theta(a) \log \frac{\pi_\theta(a)}{\pi_{\rm ref}(a)}.
\end{alignat*}
\end{definition}

$k1$ is used in aligning InstructGPT \citep{ouyang2022training}, while $k3$ is used in DeepSeek-Math \citep{shao2024deepseekmath}.

\subsubsection{Expectation Properties}

Under $\mathbb{E}_{\pi_\theta}$:
\begin{itemize}[leftmargin=2em]
    \item $\mathbb{E}_{\pi_\theta}[\hat{D}^{(k1)}] = \KL[\pi_\theta \| \pi_{\rm ref}]$. \textbf{Unbiased}, but can be negative for individual tokens.
    \item $\mathbb{E}_{\pi_\theta}[\hat{D}^{(k2)}] = \frac{1}{2}\text{Var}_{\pi_\theta}[\delta] + \frac{1}{2}(\KL)^2 \neq \KL$ in general. \textbf{Biased} as a value estimator.
    \item $\mathbb{E}_{\pi_\theta}[\hat{D}^{(k3)}] = \KL[\pi_\theta \| \pi_{\rm ref}]$ (Proposition~\ref{prop:k3_unbiased}). \textbf{Unbiased} and always $\geq 0$.
    \item $\hat{D}^{(\text{full})}$ is \textbf{exact} (no sampling involved).
\end{itemize}

\subsubsection{Gradient Properties}

The gradient of the expected KL loss under the sampling distribution $\pi_{\theta_{\rm old}}$ depends critically on the estimator choice. We analyze $\nabla_\theta \mathbb{E}_{\pi_{\theta_{\rm old}}}[\hat{D}]$ for each estimator, noting that $\pi_{\rm ref}$ is constant w.r.t.\ $\theta$.

\paragraph{k1 gradient.}
Since $\nabla_\theta \hat{D}^{(k1)} = \nabla_\theta \log \pi_\theta(o_t \mid s_t)$, the expected gradient under $\pi_\theta$ is:
\begin{align*}
    \mathbb{E}_{\pi_\theta}\!\left[\nabla_\theta \hat{D}^{(k1)}\right] = \mathbb{E}_{\pi_\theta}\!\left[\nabla_\theta \log \pi_\theta\right] = \sum_{o\in\mathcal V} \pi_\theta(o|s) \frac{\nabla_\theta \pi_\theta(o|s)}{\pi_\theta(o|s)} = \nabla_\theta \sum_{o\in\mathcal V} \pi_\theta(o|s) = 0.
\end{align*}
This is the score function identity---the expected gradient of $k1$ is zero under $\pi_\theta$, \emph{not} $\nabla_\theta \KL$. Under $\pi_{\theta_{old}}$ (without importance weighting), $\mathbb{E}_{\pi_{\theta_{\rm old}}}[\nabla_\theta \hat{D}^{(k1)}]$ gives a \textbf{biased} gradient estimate of $\nabla_\theta \KL[\pi_\theta \| \pi_{\rm ref}]$.

\paragraph{k2 gradient.}
We have $\nabla_\theta \hat{D}^{(k2)} = \delta \cdot \nabla_\theta \delta = \delta \cdot \nabla_\theta \log\pi_\theta$.

\begin{proposition}\label{prop:k2_gradient}
Under on-policy sampling, $k2$ provides an unbiased gradient estimator: $\mathbb{E}_{\pi_\theta}\!\left[\nabla_\theta \hat{D}^{(k2)}\right] = \nabla_\theta \KL[\pi_\theta \| \pi_{\rm ref}]$.
\end{proposition}
\begin{proof}
Since $\nabla_\theta \delta = \nabla_\theta \log\pi_\theta$ (the $\pi_{\rm ref}$ term vanishes):
\begin{align*}
    \nabla_\theta \KL[\pi_\theta \| \pi_{\rm ref}] &= \nabla_\theta \sum_a \pi_\theta(a)\log\frac{\pi_\theta(a)}{\pi_{\rm ref}(a)} \\
    &= \sum_a \nabla_\theta \pi_\theta(a) \cdot \log\frac{\pi_\theta(a)}{\pi_{\rm ref}(a)} + \sum_a \pi_\theta(a) \cdot \frac{\nabla_\theta \pi_\theta(a)}{\pi_\theta(a)} \\
    &= \sum_a \pi_\theta(a) \cdot \nabla_\theta\log\pi_\theta(a) \cdot \delta_a + \underbrace{\sum_a \nabla_\theta\pi_\theta(a)}_{=\,0} \\
    &= \mathbb{E}_{\pi_\theta}\!\left[\nabla_\theta\log\pi_\theta \cdot \delta\right] = \mathbb{E}_{\pi_\theta}\!\left[\nabla_\theta \hat{D}^{(k2)}\right]. \qedhere
\end{align*}
\end{proof}

Moreover, under $\pi_{\theta_{\rm old}} \approx \pi_\theta$ (e.g., early stage in PPO epochs), it remains approximately unbiased. Under off-policy sampling ($\pi_{\theta_{\rm old}} \neq \pi_\theta$), the $k2$ gradient estimator also becomes biased, though the bias is generally smaller than that of $k1$ or $k3$.

\paragraph{k3 gradient.}
We have:
\begin{align*}
    \nabla_\theta \hat{D}^{(k3)} = \nabla_\theta\!\left[\frac{1}{\alpha} + \log \alpha - 1\right] = \left(1 - \frac{1}{\alpha}\right)\nabla_\theta \log\pi_\theta.
\end{align*}
The expected gradient under $\pi_\theta$:
\begin{align*}
    \mathbb{E}_{\pi_\theta}\!\left[\nabla_\theta \hat{D}^{(k3)}\right] &= \mathbb{E}_{\pi_\theta}\!\left[\left(1 - \frac{\pi_{\rm ref}}{\pi_\theta}\right)\nabla_\theta\log\pi_\theta\right] \\
    &= \sum_a \nabla_\theta\pi_\theta(a) - \sum_a \pi_{\rm ref}(a)\frac{\nabla_\theta\pi_\theta(a)}{\pi_\theta(a)} \\
    &= 0 - \mathbb{E}_{\pi_{\rm ref}}\!\left[\frac{\nabla_\theta\pi_\theta}{\pi_\theta}\right]\\
    &=- \mathbb{E}_{\pi_{\rm ref}}\!\left[\frac{\nabla_\theta\pi_\theta}{\pi_\theta}\right].
\end{align*}
In general, this does \emph{not} equal $\nabla_\theta \KL[\pi_\theta \| \pi_{\rm ref}]$. Thus the expected gradient of $k3$ is \textbf{biased}.

\paragraph{Straight-through trick (verl's $k3+$ \citep{sheng2024hybridflow}).}
Define:
\begin{align*}
    \hat{D}^{(k3+)} = \underbrace{\hat{D}^{(k2)}}_{\text{backward}} - \text{sg}\!\left(\hat{D}^{(k2)}\right) + \text{sg}\!\left(\hat{D}^{(k3)}\right),
\end{align*}
where $\text{sg}(\cdot)$ denotes stop-gradient. This yields:
\begin{align*}
    \text{Forward value:} &\quad \hat{D}^{(k3+)} = \hat{D}^{(k3)}, \\
    \text{Backward gradient:} &\quad \nabla_\theta \hat{D}^{(k3+)} = \nabla_\theta \hat{D}^{(k2)} = \delta \cdot \nabla_\theta\log\pi_\theta.
\end{align*}
This combines the unbiased \emph{value} of $k3$ with the unbiased \emph{gradient} of $k2$.

We briefly summarize the properties of different KL estimators in Table~\ref{tab:estimators}.

\begin{table}[h]
\centering
\caption{Properties of Reverse KL estimators under $\mathbb{E}_{\pi_\theta}$. All sample-based estimators become biased under $\mathbb{E}_{\pi_{\theta_{\rm old}}}$ when $\pi_{\theta_{\rm old}} \neq \pi_\theta$.}
\label{tab:estimators}
\begin{tabular}{lcccc}
\toprule
Estimator & Value unbiased & Gradient unbiased & Non-negative \\
\midrule
$k1$ & \checkmark & $\times$ & $\times$ \\
$k2$ & $\times$ & \checkmark & \checkmark \\
$k3$ & \checkmark & $\times$ & \checkmark \\
$k3+$ & \checkmark & \checkmark & \checkmark \\
full & exact & exact & \checkmark \\
\bottomrule
\end{tabular}\
\end{table}

\subsection{Forward KL Estimators}
\label{sec:app:forward_kl}

For the forward KL formulation $\mathbb D_{\rm KL}(\pi_{\rm ref}\|\pi_\theta)$, we can similarly define KL estimators by reversing the log-ratio:
\begin{definition}[Forward KL Estimators]
    \begin{align*}
        \hat D^{(k1)}_{\rm forward}&=\log \alpha^{-1}=\log u=-\delta,\\
        \hat D^{(k2)}_{\rm forward}&=\frac{1}{2}(-\delta)^2=\frac{1}{2}\delta^2=\frac{1}{2}(\log \alpha)^2,\\
        \hat D^{(k3)}_{\rm forward}&=\alpha-\delta-1,\\
        \hat D^{\rm (full)}_{\rm forward}&=\sum_{a\in\mathcal V}\pi_{\rm ref}(a)\log\frac{\pi_{\rm ref}(a)}{\pi_\theta(a)}.
    \end{align*}
\end{definition}

\subsubsection{Expectation properties.}
Under $\mathbb E_{\pi_{\rm ref}}$:
\[
    \mathbb E_{\pi_{\rm ref}}
    \left[
        \hat D_{\rm forward}^{(k1)}
    \right]
    =
    \mathbb E_{\pi_{\rm ref}}[-\delta]
    =
    \mathbb D_{\rm KL}[\pi_{\rm ref}\|\pi_\theta].
\]
Thus $k1$ is an unbiased value estimator for the forward KL, but it can be negative for individual tokens.

For $k2$, we have
\[
    \mathbb E_{\pi_{\rm ref}}
    \left[
        \hat D_{\rm forward}^{(k2)}
    \right]
    =
    \frac12
    \mathbb E_{\pi_{\rm ref}}[\delta^2]
    =
    \frac12
    \operatorname{Var}_{\pi_{\rm ref}}[\delta]
    +
    \frac12
    \left(
        \mathbb D_{\rm KL}[\pi_{\rm ref}\|\pi_\theta]
    \right)^2
    \neq
    \mathbb D_{\rm KL}[\pi_{\rm ref}\|\pi_\theta]
\]
in general. Hence $k2$ is biased as a value estimator.

For $k3$, since $\mathbb E_{\pi_{\rm ref}}[\alpha]=1$, we obtain
\[
    \mathbb E_{\pi_{\rm ref}}
    \left[
        \hat D_{\rm forward}^{(k3)}
    \right]
    =
    \mathbb E_{\pi_{\rm ref}}[\alpha-\delta-1]
    =
    \mathbb E_{\pi_{\rm ref}}[-\delta]
    =
    \mathbb D_{\rm KL}[\pi_{\rm ref}\|\pi_\theta].
\]
Moreover, by the inequality $x-\log x-1\ge 0$ for $x>0$, $\hat D_{\rm forward}^{(k3)}\ge 0$ for every token. Therefore, $k3$ is an unbiased and nonnegative value estimator for the forward KL.

Finally, $\hat D_{\rm forward}^{(full)}$ is exact, since it explicitly sums over the whole vocabulary.

\subsubsection{Gradient properties.}
We now analyze
\[
    \nabla_\theta
    \mathbb E_{\pi_{\rm ref}}
    \left[
        \hat D_{\rm forward}
    \right],
\]
noting that $\pi_{\rm ref}$ is constant with respect to $\theta$.

\paragraph{k1 gradient.}
Since
\[
    \nabla_\theta
    \hat D_{\rm forward}^{(k1)}
    =
    \nabla_\theta(-\delta)
    =
    -\nabla_\theta\log\pi_\theta,
\]
we have
\[
    \mathbb E_{\pi_{\rm ref}}
    \left[
        \nabla_\theta
        \hat D_{\rm forward}^{(k1)}
    \right]
    =
    -
    \mathbb E_{\pi_{\rm ref}}
    \left[
        \nabla_\theta\log\pi_\theta
    \right].
\]
On the other hand,
\[
\begin{aligned}
    \nabla_\theta
    \mathbb D_{\rm KL}[\pi_{\rm ref}\|\pi_\theta]
    &=
    \nabla_\theta
    \sum_a
    \pi_{\rm ref}(a)
    \log
    \frac{\pi_{\rm ref}(a)}{\pi_\theta(a)}
    \\
    &=
    -
    \sum_a
    \pi_{\rm ref}(a)
    \nabla_\theta\log\pi_\theta(a)
    \\
    &=
    -
    \mathbb E_{\pi_{\rm ref}}
    \left[
        \nabla_\theta\log\pi_\theta
    \right].
\end{aligned}
\]
Therefore,
\[
    \mathbb E_{\pi_{\rm ref}}
    \left[
        \nabla_\theta
        \hat D_{\rm forward}^{(k1)}
    \right]
    =
    \nabla_\theta
    \mathbb D_{\rm KL}[\pi_{\rm ref}\|\pi_\theta].
\]
Thus $k1$ provides an unbiased gradient estimator for the forward KL under $\pi_{\rm ref}$ sampling.

\paragraph{k2 gradient.}
For $k2$, we have
\[
    \nabla_\theta
    \hat D_{\rm forward}^{(k2)}
    =
    \nabla_\theta
    \frac12\delta^2
    =
    \delta\cdot\nabla_\theta\log\pi_\theta.
\]
Hence
\[
    \mathbb E_{\pi_{\rm ref}}
    \left[
        \nabla_\theta
        \hat D_{\rm forward}^{(k2)}
    \right]
    =
    \mathbb E_{\pi_{\rm ref}}
    \left[
        \delta\cdot\nabla_\theta\log\pi_\theta
    \right].
\]
This is not equal to
\[
    \nabla_\theta
    \mathbb D_{\rm KL}[\pi_{\rm ref}\|\pi_\theta]
    =
    -
    \mathbb E_{\pi_{\rm ref}}
    \left[
        \nabla_\theta\log\pi_\theta
    \right]
\]
in general. Therefore, although $k2$ is a second-order approximation to the KL value near $\alpha=1$, it is generally biased as a gradient estimator for the forward KL.

\paragraph{k3 gradient.}
For $k3$, we have
\[
\begin{aligned}
    \nabla_\theta
    \hat D_{\rm forward}^{(k3)}
    &=
    \nabla_\theta(\alpha-\delta-1)
    \\
    &=
    \alpha\nabla_\theta\log\pi_\theta
    -
    \nabla_\theta\log\pi_\theta
    \\
    &=
    (\alpha-1)
    \nabla_\theta\log\pi_\theta.
\end{aligned}
\]
Taking expectation under $\pi_{\rm ref}$ gives
\[
\begin{aligned}
    \mathbb E_{\pi_{\rm ref}}
    \left[
        \nabla_\theta
        \hat D_{\rm forward}^{(k3)}
    \right]
    &=
    \mathbb E_{\pi_{\rm ref}}
    \left[
        (\alpha-1)
        \nabla_\theta\log\pi_\theta
    \right]
    \\
    &=
    \mathbb E_{\pi_{\rm ref}}
    \left[
        \alpha
        \nabla_\theta\log\pi_\theta
    \right]
    -
    \mathbb E_{\pi_{\rm ref}}
    \left[
        \nabla_\theta\log\pi_\theta
    \right]
    \\
    &=
    \mathbb E_{\pi_\theta}
    \left[
        \nabla_\theta\log\pi_\theta
    \right]
    -
    \mathbb E_{\pi_{\rm ref}}
    \left[
        \nabla_\theta\log\pi_\theta
    \right]
    \\
    &=
    0
    -
    \mathbb E_{\pi_{\rm ref}}
    \left[
        \nabla_\theta\log\pi_\theta
    \right]
    \\
    &=
    \nabla_\theta
    \mathbb D_{\rm KL}[\pi_{\rm ref}\|\pi_\theta].
\end{aligned}
\]
Thus $k3$ also provides an unbiased gradient estimator for the forward KL under $\pi_{\rm ref}$ sampling, while additionally being nonnegative and value-unbiased.

Under $\pi_{\rm ref}$ sampling, the forward KL estimators have the following properties in Table~\ref{tab:forward_kl_estimators}.

\begin{table}[h]
    \centering
    \caption{Properties for Forward KL Estimators}
    \label{tab:forward_kl_estimators}
    \begin{tabular}{lccc}
        \toprule
        \textbf{Estimator} & \textbf{Value Unbiased} & \textbf{Gradient Unbiased} & \textbf{Nonnegative} \\\midrule
        $k1$ & \checkmark & \checkmark & $\times$ \\
        $k2$ & $\times$ & $\times$ & \checkmark \\
        $k3$ & \checkmark & \checkmark & \checkmark \\
        full & exact & exact & \checkmark\\
        \bottomrule
    \end{tabular}
\end{table}

\section{Implicit Arcsin Objective of Binary GRPO}
\label{app:arcsin-allocation}

This appendix proves Proposition~\ref{prop:grpo_arcsin_objective}. We use the same notation as the main text. Fix a prompt $x$, let $a\sim\pi_\theta(\cdot|x)$ be a sampled response, and assume a binary verifier reward $r(x,a)\in\{0,1\}$. Define
\[
    p_\theta(x)
    :=
    \mathbb E_{a\sim\pi_\theta(\cdot|x)}[r(x,a)].
\]
We analyze the population version of GRPO, replacing the empirical group mean and standard deviation by
\[
    \mu(x)=p_\theta(x),
    \qquad
    \sigma(x)=\sqrt{p_\theta(x)(1-p_\theta(x))},
\]
for $p_\theta(x)\in(0,1)$. The corresponding population GRPO advantage is
\[
    A^{\rm GRPO}(x,a)
    =
    \frac{r(x,a)-p_\theta(x)}
    {\sqrt{p_\theta(x)(1-p_\theta(x))}}.
\]

\begin{proof}[Proof of Proposition~\ref{prop:grpo_arcsin_objective}]
By the likelihood-ratio identity,
\[
    \nabla_\theta p_\theta(x)
    =
    \nabla_\theta
    \mathbb E_{a\sim\pi_\theta(\cdot|x)}[r(x,a)]
    =
    \mathbb E_{a\sim\pi_\theta(\cdot|x)}
    \left[
        r(x,a)\nabla_\theta\log\pi_\theta(a|x)
    \right].
\]
Also,
\[
    \mathbb E_{a\sim\pi_\theta(\cdot|x)}
    [\nabla_\theta\log\pi_\theta(a|x)]
    =
    0.
\]
Therefore,
\begin{align*}
    &\mathbb E_{a\sim\pi_\theta(\cdot|x)}
    \left[
        A^{\rm GRPO}(x,a)\nabla_\theta\log\pi_\theta(a|x)
    \right] \\
    &\qquad =
    \frac{
    \mathbb E_{a\sim\pi_\theta(\cdot|x)}
    [
        (r(x,a)-p_\theta(x))
        \nabla_\theta\log\pi_\theta(a|x)
    ]}
    {\sqrt{p_\theta(x)(1-p_\theta(x))}} \\
    &\qquad =
    \frac{\nabla_\theta p_\theta(x)}
    {\sqrt{p_\theta(x)(1-p_\theta(x))}}.
\end{align*}
Finally,
\[
    \frac{d}{dp}\,2\arcsin\sqrt p
    =
    \frac{1}{\sqrt{p(1-p)}},
\]
so the last display equals
\[
    \nabla_\theta
    \left[
        2\arcsin\sqrt{p_\theta(x)}
    \right].
\]
\end{proof}

\begin{remark}[Finite groups and degenerate cases]
The proof uses population group statistics. In finite groups, GRPO replaces $p_\theta(x)$ and $\sqrt{p_\theta(x)(1-p_\theta(x))}$ by their empirical estimates. This introduces finite-sample noise and degenerate groups when all sampled responses have the same binary reward, in which case $\sigma_G=0$ and implementations typically mask or skip the prompt. The proposition should therefore be read as the population objective followed by the GRPO estimator.
\end{remark}

\begin{remark}[Variance-stabilizing intuition]
For a Bernoulli reward, $A^{\rm GRPO}$ is the standardized reward and satisfies
\[
    \mathbb E[A^{\rm GRPO}]=0,
    \qquad
    \mathbb E[(A^{\rm GRPO})^2]=1.
\]
Thus group normalization removes the direct $p_\theta(x)(1-p_\theta(x))$ reward-variance factor. The transform $2\arcsin\sqrt p$ is the corresponding variance-stabilizing transform for Bernoulli means.
\end{remark}

\begin{remark}[Score-of-utility view; cf.\ Corollary~\ref{cor:grpo_arcsin_score}]
The proof above also establishes Corollary~\ref{cor:grpo_arcsin_score}: the standardized GRPO advantage is exactly $\ell'(p_\theta(x))(r(x,a)-p_\theta(x))$ for $\ell(p)=2\arcsin\sqrt p$. The non-linearity of $\ell$ is the structural reason that any reward-side modification (in particular, KL-in-reward) is rescaled by $\ell'(p)$ together with $r$, while a token-level regularizer added \emph{after} normalization is not — see Proposition~\ref{prop:kl_in_reward_distortion} for the resulting non-commutativity of $\ell$ with KL shaping.
\end{remark}

\section{Missing Proofs}
\label{sec:app:proofs}

\subsection{Proof Sketch of Proposition~\ref{prop:kl_in_reward_distortion}}
\label{sec:app:proofs:kl_in_reward_distortion}

We give a short structural argument; the key observation is that the population gradient driven by the GRPO advantage on a KL-shaped reward targets the score of the variance-stabilizing transform of the shaped reward, not the sum of two independent gradients.

Under binary $r$ and the population GRPO advantage from Proposition~\ref{prop:grpo_arcsin_objective}, the expected policy gradient equals the gradient of the implicit utility $\ell(p_\theta(x))=2\arcsin\sqrt{p_\theta(x)}$ (Corollary~\ref{cor:grpo_arcsin_score}). Now replace $r_i$ by the KL-shaped trajectory reward $\tilde r_i=r_i-\beta\sum_j\delta_{i,j}$, with population statistics $\tilde\mu(x)=\mathbb E[\tilde r]=p_\theta(x)-\beta\,K(\theta)$ and $\tilde\sigma(x)=\mathrm{Std}[\tilde r]$, where $K(\theta)=\mathbb E_{a\sim\pi_\theta}\sum_j\delta_j$ abbreviates the prompt-level KL.

The standardized advantage $\hat A^{\rm KLR}(x,a)=(\tilde r(x,a)-\tilde\mu(x))/\tilde\sigma(x)$ drives, in expectation,
\[
    \mathbb E_{a\sim\pi_\theta}\!\left[\hat A^{\rm KLR}(x,a)\,\nabla_\theta\log\pi_\theta(a|x)\right]
    =
    \frac{1}{\tilde\sigma(x)}\,\nabla_\theta\bigl(p_\theta(x)-\beta\,K(\theta)\bigr),
\]
by the same chain-rule identity used in Proposition~\ref{prop:grpo_arcsin_objective}. Two structural consequences follow.

\textbf{(i)} The factor $1/\tilde\sigma(x)$ multiplies \emph{both} $\nabla_\theta p_\theta$ and $\beta\,\nabla_\theta K(\theta)$. The desired objective $\ell(p_\theta(x))-\beta\,K(\theta)$ has its KL gradient unweighted by $\ell'$, so the KL-in-reward update misweights the regularization signal by exactly $\ell'(p_\theta(x))=1/\sqrt{p_\theta(x)(1-p_\theta(x))}$, which varies across prompts and over training. The two gradients agree only when $\ell'\equiv\text{const}$, i.e., when $\ell$ is affine.

\textbf{(ii)} The denominator $\tilde\sigma(x)$ itself depends on $\beta$: for binary $r$,
\[
    \mathrm{Var}[\tilde r]
    =
    p_\theta(1-p_\theta)
    -
    2\beta\,\mathrm{Cov}[r,D]
    +
    \beta^2\,\mathrm{Var}[D],
\]
where $D=\sum_j\delta_j$, so the implicit utility carries a non-trivial $\beta$-dependence that does not factor as ``$\ell$ minus regularizer.''

Together (i) and (ii) imply that the population objective targeted by $\hat A^{\rm KLR}$ reduces to $\ell(\mathbb E[r])-\beta\,\mathbb E[\KL]$ only in the affine case $\ell(p)=ap+b$, which fails for GRPO's $\ell(p)=2\arcsin\sqrt p$. \qed

\subsection{KL Estimator Value Unbiasedness}
\label{sec:app:proofs:k3_unbiased}

\begin{proof}[Proof of Proposition~\ref{prop:k3_unbiased}]
Fix a state $s$ and write $u(a)=\pi_{\rm ref}(a\mid s)/\pi_\theta(a\mid s)$. Under on-policy sampling $a\sim\pi_\theta(\cdot\mid s)$,
\begin{align*}
    \mathbb E_{\pi_\theta}\!\left[\hat D^{(k3)}\right]
    &=
    \sum_{a\in\mathcal V}
    \pi_\theta(a\mid s)
    \left[
    \frac{\pi_{\rm ref}(a\mid s)}{\pi_\theta(a\mid s)}
    +
    \log\frac{\pi_\theta(a\mid s)}{\pi_{\rm ref}(a\mid s)}
    -1
    \right] \\
    &=
    \sum_{a\in\mathcal V}\pi_{\rm ref}(a\mid s)
    +
    \sum_{a\in\mathcal V}\pi_\theta(a\mid s)
    \log\frac{\pi_\theta(a\mid s)}{\pi_{\rm ref}(a\mid s)}
    -
    \sum_{a\in\mathcal V}\pi_\theta(a\mid s) \\
    &=
    \KL[\pi_\theta(\cdot\mid s)\|\pi_{\rm ref}(\cdot\mid s)].
\end{align*}
\end{proof}

\subsection{On-Policy GRPO with KL Regularization}
\label{sec:app:proofs:future_kl}

\subsubsection{Derivation of the Regularized Policy-Gradient Weight}
\label{sec:app:proofs:kl_regularized_pg_weight}
\begin{proof}
We derive the policy-gradient weight in Eq.~\eqref{eq:kl_regularized_pg_weight}.
Recall the objective function,
    \begin{align*}
Q(\theta)
&=
\mathbb E_x \ell\!\left(\mathbb E_{a\sim\pi_\theta} r(x,a)\right)
-
\beta\cdot \mathbb E_x \mathbb E_{a\sim\pi_\theta}
\sum_{i=1}^T
f\!\left(
    \frac{\pi_{\rm ref}(a_i\mid s_i)}
         {\pi_\theta(a_i\mid s_i)}
\right).
\end{align*}

For a fixed prompt $x$, denote
\begin{align*}
\bar r_\theta(x)
&:=
\mathbb E_{a\sim\pi_\theta} r(x,a), \\
u_i
&:=
\frac{\pi_{\rm ref}(a_i\mid s_i)}
     {\pi_\theta(a_i\mid s_i)}, \\
\pi_\theta(a\mid x)
&=
\prod_{i=1}^T \pi_\theta(a_i\mid s_i).
\end{align*}

The objective is
\begin{align*}
Q(\theta)
&=
\mathbb E_x \ell(\bar r_\theta(x))
-
\beta
\mathbb E_x
\mathbb E_{a\sim\pi_\theta}
\sum_{i=1}^T f(u_i).
\end{align*}

Then
\begin{align*}
\nabla_\theta Q(\theta)
&=
\nabla_\theta
\mathbb E_x \ell(\bar r_\theta(x))
-
\beta
\nabla_\theta
\mathbb E_x
\mathbb E_{a\sim\pi_\theta}
\sum_{j=1}^T f(u_j).
\end{align*}

First, consider the reward term. By the chain rule,
\begin{align*}
\nabla_\theta \ell(\bar r_\theta(x))
&=
\ell'(\bar r_\theta(x))
\nabla_\theta \bar r_\theta(x) \\
&=
\ell'(\bar r_\theta(x))
\nabla_\theta
\mathbb E_{a\sim\pi_\theta} r(x,a).
\end{align*}

Using the likelihood-ratio identity,
\begin{align*}
\nabla_\theta
\mathbb E_{a\sim\pi_\theta} r(x,a)
&=
\nabla_\theta
\sum_a \pi_\theta(a\mid x) r(x,a) \\
&=
\sum_a
\nabla_\theta \pi_\theta(a\mid x) r(x,a) \\
&=
\sum_a
\pi_\theta(a\mid x)
\nabla_\theta \log \pi_\theta(a\mid x)
r(x,a) \\
&=
\mathbb E_{a\sim\pi_\theta}
\left[
    r(x,a)
    \nabla_\theta \log \pi_\theta(a\mid x)
\right].
\end{align*}

Since
\begin{align*}
\log \pi_\theta(a\mid x)
&=
\sum_{i=1}^T \log \pi_\theta(a_i\mid s_i),
\end{align*}
we have
\begin{align*}
\nabla_\theta
\mathbb E_{a\sim\pi_\theta} r(x,a)
&=
\mathbb E_{a\sim\pi_\theta}
\left[
    r(x,a)
    \sum_{i=1}^T
    \nabla_\theta\log\pi_\theta(a_i\mid s_i)
\right] \\
&=
\mathbb E_{a\sim\pi_\theta}
\sum_{i=1}^T
r(x,a)
\nabla_\theta\log\pi_\theta(a_i\mid s_i).
\end{align*}

Therefore,
\begin{align*}
\nabla_\theta \ell(\bar r_\theta(x))
&=
\mathbb E_{a\sim\pi_\theta}
\sum_{i=1}^T
\ell'(\bar r_\theta(x))r(x,a)
\nabla_\theta\log\pi_\theta(a_i\mid s_i).
\end{align*}

For any baseline $b(s_i)$,
\begin{align*}
&\mathbb E_{a_i\sim\pi_\theta(\cdot\mid s_i)}
\left[
    b(s_i)
    \nabla_\theta\log\pi_\theta(a_i\mid s_i)
\right]\\
=\ &
b(s_i)
\sum_{a_i}
\pi_\theta(a_i\mid s_i)
\nabla_\theta\log\pi_\theta(a_i\mid s_i) \\
=\ &
b(s_i)
\sum_{a_i}
\nabla_\theta \pi_\theta(a_i\mid s_i) \\
=\ &
b(s_i)
\nabla_\theta
\sum_{a_i}
\pi_\theta(a_i\mid s_i) \\
=\ &0.
\end{align*}

Thus the reward contribution can be written as
\begin{align*}
\nabla_\theta \ell(\bar r_\theta(x))
&=
\mathbb E_{a\sim\pi_\theta}
\sum_{i=1}^T
A_i
\nabla_\theta\log\pi_\theta(a_i\mid s_i),
\end{align*}
where
\begin{align*}
A_i
&:=
\ell'(\bar r_\theta(x))r(x,a)-b(s_i) \\
&=
\ell'\!\left(\mathbb E_{a\sim\pi_\theta}r(x,a)\right)r(x,a)-b(s_i).
\end{align*}

Next, consider the regularization term. Define
\begin{align*}
G(\theta;x)
&:=
\mathbb E_{a\sim\pi_\theta}
\sum_{j=1}^T f(u_j).
\end{align*}

Then
\begin{align*}
\nabla_\theta G(\theta;x)
&=
\nabla_\theta
\sum_a
\pi_\theta(a\mid x)
\sum_{j=1}^T f(u_j) \\
&=
\sum_a
\nabla_\theta \pi_\theta(a\mid x)
\sum_{j=1}^T f(u_j)
+
\sum_a
\pi_\theta(a\mid x)
\nabla_\theta
\sum_{j=1}^T f(u_j) \\
&=
\mathbb E_{a\sim\pi_\theta}
\left[
    \left(\sum_{j=1}^T f(u_j)\right)
    \nabla_\theta\log\pi_\theta(a\mid x)
    +
    \sum_{j=1}^T
    \nabla_\theta f(u_j)
\right].
\end{align*}

For each $j$,
\begin{align*}
u_j
&=
\frac{\pi_{\rm ref}(a_j\mid s_j)}
     {\pi_\theta(a_j\mid s_j)}.
\end{align*}

The derivative below is the direct derivative of the sampled token penalty
conditional on the realized prefix $s_j$; the dependence of later prefixes on
earlier actions is accounted for by the score-function term above. Since
$\pi_{\rm ref}$ is fixed,
\begin{align*}
\nabla_\theta u_j
&=
\nabla_\theta
\frac{\pi_{\rm ref}(a_j\mid s_j)}
     {\pi_\theta(a_j\mid s_j)} \\
&=
-\frac{\pi_{\rm ref}(a_j\mid s_j)}
       {\pi_\theta(a_j\mid s_j)^2}
\nabla_\theta \pi_\theta(a_j\mid s_j) \\
&=
-\frac{\pi_{\rm ref}(a_j\mid s_j)}
       {\pi_\theta(a_j\mid s_j)}
\nabla_\theta\log\pi_\theta(a_j\mid s_j) \\
&=
-u_j
\nabla_\theta\log\pi_\theta(a_j\mid s_j).
\end{align*}

Hence
\begin{align*}
\nabla_\theta f(u_j)
&=
f'(u_j)\nabla_\theta u_j \\
&=
-f'(u_j)u_j
\nabla_\theta\log\pi_\theta(a_j\mid s_j).
\end{align*}

Substituting this into $\nabla_\theta G(\theta;x)$ gives
\begin{align*}
\nabla_\theta G(\theta;x)
&=
\mathbb E_{a\sim\pi_\theta}
\left[
    \left(\sum_{j=1}^T f(u_j)\right)
    \sum_{i=1}^T
    \nabla_\theta\log\pi_\theta(a_i\mid s_i)
    -
    \sum_{j=1}^T
    f'(u_j)u_j
    \nabla_\theta\log\pi_\theta(a_j\mid s_j)
\right] \\
&=
\mathbb E_{a\sim\pi_\theta}
\sum_{i=1}^T
\left(
    \sum_{j=1}^T f(u_j)
    -
    f'(u_i)u_i
\right)
\nabla_\theta\log\pi_\theta(a_i\mid s_i).
\end{align*}

Now we convert the first summation into its causal reward-to-go form. For
$j<i$, the quantity $f(u_j)$ is measurable with respect to the history
$s_i=(x,a_{<i})$. Therefore,
\begin{align*}
&\mathbb E_{a\sim\pi_\theta}
\left[
    f(u_j)
    \nabla_\theta\log\pi_\theta(a_i\mid s_i)
\right]\\
=\ &
\mathbb E_{s_i}
\left[
    f(u_j)
    \mathbb E_{a_i\sim\pi_\theta(\cdot\mid s_i)}
    \left[
        \nabla_\theta\log\pi_\theta(a_i\mid s_i)
    \right]
\right] \\
=\ &
\mathbb E_{s_i}
\left[
    f(u_j)
    \sum_{a_i}
    \pi_\theta(a_i\mid s_i)
    \nabla_\theta\log\pi_\theta(a_i\mid s_i)
\right] \\
=\ &
\mathbb E_{s_i}
\left[
    f(u_j)
    \sum_{a_i}
    \nabla_\theta\pi_\theta(a_i\mid s_i)
\right] \\
=\ &
\mathbb E_{s_i}
\left[
    f(u_j)
    \nabla_\theta
    \sum_{a_i}
    \pi_\theta(a_i\mid s_i)
\right] \\
=\ &0.
\end{align*}

Thus the past terms $j<i$ can be dropped from the score-function coefficient,
and we obtain
\begin{align*}
\nabla_\theta G(\theta;x)
&=
\mathbb E_{a\sim\pi_\theta}
\sum_{i=1}^T
\left(
    \sum_{j\ge i} f(u_j)
    -
    f'(u_i)u_i
\right)
\nabla_\theta\log\pi_\theta(a_i\mid s_i).
\end{align*}

Finally, since
\begin{align*}
Q(\theta)
&=
\mathbb E_x \ell(\bar r_\theta(x))
-
\beta \mathbb E_x G(\theta;x),
\end{align*}
we have
\begin{align*}
\nabla_\theta Q(\theta)
&=
\mathbb E_x
\mathbb E_{a\sim\pi_\theta}
\sum_{i=1}^T
\left[
    A_i
    -
    \beta
    \left(
        \sum_{j\ge i} f(u_j)
        -
        f'(u_i)u_i
    \right)
\right]
\nabla_\theta\log\pi_\theta(a_i\mid s_i).
\end{align*}

Therefore, defining
\begin{align*}
w_i
&:=
\operatorname{stopgrad}
\left(
    A_i
    -
    \beta
    \left(
        \sum_{j\ge i} f(u_j)
        -
        f'(u_i)u_i
    \right)
\right),
\end{align*}
we get
\begin{align*}
\nabla_\theta Q(\theta)
&=
\mathbb E_x
\mathbb E_{a\sim\pi_\theta}
\sum_{i=1}^T
w_i
\nabla_\theta\log\pi_\theta(a_i\mid s_i).
\end{align*}
\end{proof}

\subsection{KL in Reward from the Perspective of Optimality}
\label{sec:app:proofs:prl}

\subsubsection{Proof of Theorem~\ref{prop:prl_optimal}}
\label{sec:app:proofs:prl_optimal}

\begin{proof}
Consider for a fixed prompt $x\in\cV^*$, and our optimization objective is:

\[\begin{aligned}
     & \max_{\pi(\cdot | x)}\{\mathbb{E}_{a \sim \pi(\cdot | x)} [r(x, a)] - \beta\KL(\pi \| \pi_{\rm ref})\} \\
    =& \max_{\pi(\cdot | x)} \Bigg\{ \sum_{a \in \cV^*} \pi(a | x) \cdot \left(r(x, a) - \beta \ln\left(\frac{\pi(a | x)}{\pi_{\rm ref}(a | x)}\right)\right) \Bigg\},
\end{aligned}\]
under the condition $\sum_{a \in \cV^*} \pi(a | x) = 1$.

Thus, we construct the Lagrangian of the above optimization problem,
\[
\mathcal{L}(\pi(\cdot|x), \lambda) = \sum_{a \in \cV^*} \pi(a | x) \cdot \left(r(x, a) - \beta \ln\left(\frac{\pi(a | x)}{\pi_{\rm ref}(a | x)}\right)\right) - \lambda \Big(1 - \sum_{a \in \cV^*}\pi(a | x)\Big).
\]
Setting the partial derivative of $\cL(\pi,\lambda)$ with respect to $\pi$ equal to 0,
\[\begin{aligned}
\frac{\partial}{\partial \pi(a | x)} \mathcal{L(\pi, \lambda)} = 0,
\end{aligned}
\]
we have
\begin{align*} 
&\frac{\partial}{\partial \pi(a | x)}\left[ \sum_{a \in \cV^*} \pi(a | x) \cdot \left(r(x, a) - \beta \ln\left(\frac{\pi(a | x)}{\pi_{\rm ref}(a | x)}\right)\right) - \lambda \left(1 - \sum_{a \in \cV^*}\pi(a | x)\right)\right] = 0\\
\Leftrightarrow &\ r(x, a) - \beta\left(\ln\left(\frac{\pi(a | x)}{\pi_{\rm ref}(a | x)}\right) + 1\right) - \lambda = 0\\
\Leftrightarrow&\ \frac 1\beta \cdot r(x, a) - \frac 1\beta \cdot \lambda - 1 = \ln\left(\pi(a | x)\right) - \ln\left(\pi_{\rm ref}(a | x)\right)\\
\Leftrightarrow&\ \ln \pi(a | x) = \ln\pi_{\rm ref}(a | x) + \frac 1\beta \cdot r(x, a) - \frac 1\beta \cdot \lambda - 1\\
\Leftrightarrow&\ \pi(a | x) = \pi_{\rm ref}(a | x) \exp\left(\frac 1\beta \cdot r(x, a) - \frac 1\beta \lambda - 1\right) \propto \pi_{\rm ref}(a | x) \exp\Big(\frac 1\beta r(x, a)\Big).
\end{align*}
Therefore, we complete the proof.
\end{proof}

\subsubsection{Proof of Theorem~\ref{prop:prl_process}}
\label{sec:app:proofs:prl_process_reward}

\begin{proof}
From Proposition~\ref{prop:prl_constant}, under optimal policy $\pi^*$ and reward $r$:
\[C = r(x, a) - \beta\ln\frac{\pi^*(a | x)}{\pi_{\rm ref}(a | x)}.\]
Given the auto-regressive generation property of language modeling, we have:
\[
\ln\frac{\pi^*(a | x)}{\pi_{\rm ref}(a | x)} = \ln\frac{\prod_{j = 1}^T \pi^*(a_j | x, a^{(j - 1)})}{\prod_{j = 1}^T \pi_{\rm ref}(a_j | x, a^{(j - 1)})} = \sum_{j = 1}^T \ln\frac{\pi^*(a_j | x, a^{(j - 1)})}{\pi_{\rm ref}(a_j | x, a^{(j - 1)})}.
\]
Therefore, we have:
\[
C = r(x, a) - \beta \sum_{j = 1}^t \ln\left(\frac{\pi^*(a_j | x, a^{(j - 1)})}{\pi_{\rm ref}(a_j | x, a^{(j - 1)})}\right) - \beta\sum_{j = t + 1}^T \ln\left(\frac{\pi^*(a_j | x, a^{(j - 1)})}{\pi_{\rm ref}(a_j | x, a^{(j - 1)})}\right)
\]
Therefore,
\[\begin{aligned}
     & r_t(x, a^{(t)}) \\
    =\ & \E_{a^{(-t)} \sim \pi^*(\cdot | x, a^{(t)})}
\left[ r(x, a) - \beta\sum_{j = t + 1}^T \ln\left(\frac{\pi^*(a_j | x, a^{(j - 1)})}{\pi_{\rm ref}(a_j | x, a^{(j - 1)})}\right)\right] \\
    =\ & \E_{a^{(-t)} \sim \pi^*(\cdot | x, a^{(t)})} \left[ C + \beta \sum_{j = 1}^t \ln\left(\frac{\pi^*(a_j | x, a^{(j - 1)})}{\pi_{\rm ref}(a_j | x, a^{(j - 1)})}\right)\right] \\
    =\ & C',
\end{aligned}\]
for some constant $C'\in\mathbb R$, where the final equation holds due to the fact that $a^{(t)}$ is fixed given the conditional distribution $\pi^*(\cdot|x,a^{(t)})$. Thus, the process reward $r_t(x, a^{(t)})$ should be a constant no matter what the future path is under the optimal policy $\pi^*$. Therefore, given the whole reasoning trajectory $a=(a^1,\cdots,a^T)$, the universal equality holds:
\[r_t(x,a^{(t)})
= r(x,a) - \beta\sum_{j=t+1}^T \ln \frac{\pi^*(a_j|x,a^{(j-1)})}{\pi_{\rm ref}(a_j|x,a^{(j-1)})} .\]
\end{proof}

\section{Algorithmic Details for \method{}}
\label{sec:app:algorithm}

\begin{algorithm}[t]
\caption{Future-KL Regularized Policy Optimization (\method{})}
\label{alg:frpo}
\small
\begin{algorithmic}[1]
\Require Policy $\pi_\theta$, reference policy $\pi_{\rm ref}$, rollout batch $\mathcal B$, mini-batch size $B_{\rm mini}$, group size $G$, KL coefficient $\beta$, PPO clip $\epsilon$
\For{each training iteration}
    \State Set rollout policy $\pi_{\theta_{\rm old}}\leftarrow \pi_\theta$
    \For{each prompt $x\in\mathcal B$}
        \State Sample responses $\{a_i\}_{i=1}^G\sim \pi_{\theta_{\rm old}}(\cdot|x)$
        \State Evaluate rewards $r_i=r(x,a_i)$ and token masks $m_{i,t}$
        \State Compute $\mu_G=\frac{1}{G}\sum_i r_i$ and $\sigma_G=\sqrt{\frac{1}{G}\sum_i(r_i-\mu_G)^2}$
        \State $\hat A_i^{\rm GRPO}\gets (r_i-\mu_G)/(\sigma_G+\varepsilon_{\rm std})$
        \For{each response token $t$ of response $i$}
            \State $\delta_{i,t}\gets \log\pi_{\theta}(a_{i,t}|x,a_{i,<t})-\log\pi_{\rm ref}(a_{i,t}|x,a_{i,<t})$ \Comment{Recompute the probability} w.r.t. $\pi_\theta$
            \State $S_{i,t}\gets \sum_{j=t}^{T_i}m_{i,j}\delta_{i,j}$ \Comment{KL-to-go from current token}
            \State $\hat A_{i,t}^{\method{}}\gets \operatorname{sg}(\hat A_i^{\rm GRPO}-\beta S_{i,t})$
        \EndFor
    \EndFor
    \State Split the rollout batch into actor mini-batches of size $B_{\rm mini}$
    \State Update $\theta$ once per mini-batch using PPO with
    \[
    L^{\rm PPO}_{i,t}(\theta)=
    \min\!\left(
    \rho_{i,t}\hat A_{i,t}^{\method{}},
    \operatorname{clip}(\rho_{i,t},1-\epsilon,1+\epsilon)\hat A_{i,t}^{\method{}}
    \right),
    \]
    \Statex \hspace{\algorithmicindent}where
    $\rho_{i,t}=\pi_\theta(a_{i,t}|x,a_{i,<t})/
    \pi_{\theta_{\rm old}}(a_{i,t}|x,a_{i,<t})$.
\EndFor
\end{algorithmic}
\end{algorithm}

\paragraph{Implementation notes.}
The default future term uses the sampled log-ratio ($k1$) estimator. A $k3$
variant replaces $\delta_{i,j}$ in $S_{i,t}$ by
$-\log u_{i,j}+u_{i,j}-1$, where
$u_{i,j}=\pi_{\rm ref}(a_{i,j}|s_{i,j})/\pi_{\theta_{\rm old}}(a_{i,j}|s_{i,j})$.
All sums are masked over response tokens and exclude prompt and padding tokens;
the default FRPO advantage includes the current response token in the KL-to-go.
With multiple PPO epochs, the future-KL term can either remain fixed from rollout
log-probabilities or be recomputed under the current policy at each actor update.
In our experiments, PPO epochs are set to $1$: a train batch is rolled out once
for efficiency and then split into actor mini-batches, each used for a single
update. Refer to Algorithm~\ref{alg:frpo} for details.

\section{Discussion about Future KL}
\label{sec:app:future_kl}

\begin{table}[h]
\centering\small
\caption{Where this paper sits among KL-regularized policy-gradient analyses. Linear $\ell$ allows reward and regularizer to fuse into a single augmented reward; under GRPO's non-linear $\ell(p)=2\arcsin\sqrt p$ they cannot, and the right insertion point for the future-KL correction is on the regularizer side of the $\ell'$-asymmetry.}
\label{tab:regimes}
\resizebox{\textwidth}{!}{
\begin{tabular}{lcccc}
\toprule
Setting & Prompt objective & Reward+KL fuseable? & Future-reg in PG? & Insertion point \\
\midrule
MaxEnt RL / soft-Q & $\mathbb E_\pi[r]-\beta\,\KL$ & yes (augmented reward) & yes (textbook) & either \\
PPO + KL-in-reward (+GAE) & $\mathbb E_\pi[r]-\beta\,\KL$ & yes & yes (via value) & reward shaping \\
\textbf{GRPO (this paper)} & $\boldsymbol{\ell(\mathbb E_\pi[r])-\beta\,\KL}$ & \textbf{no} (Prop.~\ref{prop:kl_in_reward_distortion}) & \textbf{yes (Thm.~\ref{thm:f_divergence_pg})} & \textbf{after advantage} \\
\bottomrule
\end{tabular}}
\end{table}

\subsection{Why Not Put KL into the Reward Before Normalization?}
\label{sec:kl_reward_normalization}

Section~\ref{sec:frpo_optimality} uses a decoupled update: compute the \gls{grpo} outcome advantage first, then add a token-wise future-KL correction. There are three natural ways to combine future KL with GRPO-style normalization.

\begin{enumerate}[leftmargin=2em, label=(\alph*)]
    \item \textbf{Decoupled:} Compute GRPO advantage from the original outcome reward, then add the future-KL correction:
    \begin{align*}
        \hat{A}_{i,t} = \frac{r_i - \mu_G}{\sigma_G} - \beta\cdot \sum_{j=t}^T \delta_{i,j}.
    \end{align*}
    
    \item \textbf{Coupled, trajectory-level:} Augment the outcome reward with the \emph{full} trajectory KL, then apply group normalization:
    \begin{align*}
        \tilde{r}_i = r_i - \beta\cdot\sum_{j=1}^T \delta_{i,j}, \qquad \tilde{A}_i = \frac{\tilde{r}_i - \tilde{\mu}_G}{\tilde{\sigma}_G}.
    \end{align*}
    This yields a \emph{uniform} advantage for all tokens in response $i$---the per-token differentiation from future KL is lost entirely. It also changes the group mean and standard deviation, so the regularizer affects not only policy drift but also the scale and ordering used to construct the GRPO advantage.
    
    \item \textbf{Coupled, step-level:} Use the step-dependent process reward $r_{i,t}$ and normalize across the group at each step:
    \begin{align*}
        \hat{A}_{i,t} = \frac{r_{i,t} - \mu_{G,t}}{\sigma_{G,t}}, \qquad \mu_{G,t} = \frac{1}{G}\sum_{k=1}^G r_{k,t}, \quad \sigma_{G,t} = \text{std}_k(r_{k,t}).
    \end{align*}
    In this strategy, $r_{i,t}$ uses token-level KL instead of trajectory-level KL, leading to possibly higher variance. In addition, trajectories have different lengths, so at large $t$ only a subset of trajectories contributes, further degrading the estimates.
\end{enumerate}

\paragraph{Analysis.} Strategy (b) defeats the purpose of process rewards: by collapsing the future KL into a single trajectory-level scalar, all tokens share the same advantage, and the fine-grained credit assignment from Theorem~\ref{prop:prl_process} is discarded. Moreover, as formalized in Proposition~\ref{prop:kl_in_reward_distortion}, modifying the reward before group normalization is structurally incompatible with the non-linear utility $\ell(p)=2\arcsin\sqrt p$: the modified standard deviation $\tilde{\sigma}_G \neq \sqrt{\hat{p}(1-\hat{p})}$ breaks the variance-stabilizing property that makes the GRPO gradient the score of $\ell$, and the failure does not vanish with tuning because $\ell$ is not affine. Strategy (c) is closer to the process-reward interpretation, but suffers from statistical instability: with small groups, the per-step normalization constants $\mu_{G,t}$ and $\sigma_{G,t}$ are noisy estimates, and variable trajectory lengths mean that for large $t$, very few trajectories contribute to the statistics. This can lead to erratic advantage values in the later portion of responses.

Strategy (a), used by \method{}, avoids both problems by \textbf{separating the two roles} of the advantage:
\begin{itemize}[leftmargin=2em]
    \item \emph{Trajectory ranking} is handled by $\hat{A}_i^{\rm GRPO}$, which benefits from stable group normalization on outcome rewards. Crucially, since the original reward enters the normalization unchanged, the implicit $2\arcsin(\sqrt{p})$ objective is preserved.
    \item \emph{Token differentiation} is handled by the future KL term $-\beta \cdot\sum_{j=t}^T\delta_{i,j}$, which operates within each trajectory and does not require cross-trajectory normalization.
\end{itemize}
\section{More Experiment Details}
\label{sec:app:experiment}
We show the experiment configurations in Table~\ref{tab:baseline_hyperparameters} and Table~\ref{tab:hyperparameters_ours}.
We use vLLM \citep{kwon2023efficient} as the inference backend to speedup rollout. Most experiments are conducted on 4$\times$8 NVIDIA H20 GPU nodes.

\begin{table}[h]
    \centering
    \caption{Hyperparameters for Baselines.}
    \label{tab:baseline_hyperparameters}
    \resizebox{\textwidth}{!}{
    \begin{tabular}{lccc}
    \toprule
        \textbf{Hyperparameters} & \textbf{Qwen3-1.7B-Base} & \textbf{Qwen3-4B-Base} & \textbf{Qwen3-30B-A3B-Base} \\\midrule
        Learning Rate & 1e-6 & 1e-6 & 1e-6 \\
        PPO Epochs & 1 & 1 & 1 \\
        Max Prompt Length & 2048 & 2048 & 2048 \\
        Max Response Length & 8192 & 8192 & 8192 \\
        Train Batch Size & 256 & 256 & 256 \\   
        PPO Mini Batch Size & 64 & 64 & 64 \\
        Rollout Temperature & 1.0 & 1.0 & 1.0\\
        Group Size & 5 & 5 & 5 \\
    \bottomrule
    \end{tabular}}
\end{table}

\begin{table}[h]
    \centering
    \caption{Hyperparameters for \method{}.}
    \label{tab:hyperparameters_ours}
    \begin{tabular}{lccc}
    \toprule
        \textbf{Hyperparameters} & \textbf{Qwen3-1.7B-Base} & \textbf{Qwen3-4B-Base} & \textbf{Qwen3-30B-A3B-Base} \\\midrule
         Learning Rate & 1e-6 & 1e-6 & 1e-6 \\
        PPO Epochs & 1 & 1 & 1 \\
        Max Prompt Length & 2048 & 2048 & 2048 \\
        Max Response Length & 8192 & 8192 & 8192 \\
        Train Batch Size & 256 & 256 & 256 \\   
        PPO Mini Batch Size & 64 & 64 & 64 \\
        Rollout Temperature & 1.0 & 1.0 & 1.0\\
        Group Size & 5 & 5 & 5 \\
        $\beta$ & 0.01 & 0.01 & 0.01\\
        \bottomrule
    \end{tabular}
\end{table}

For math reasoning, we add an additional prompt behind each problem, as shown in \ref{math_prompt}.
\begin{mybox}[math_prompt]{Prompt for math reasoning}
\{problem\}
\\\\
Let's think step by step and output the final answer within \textbackslash boxed\{\}.
\end{mybox}

\subsection{Experiments on Qwen3-4B-Base}

\begin{figure}[h]
    \centering
    \includegraphics[width=\linewidth]{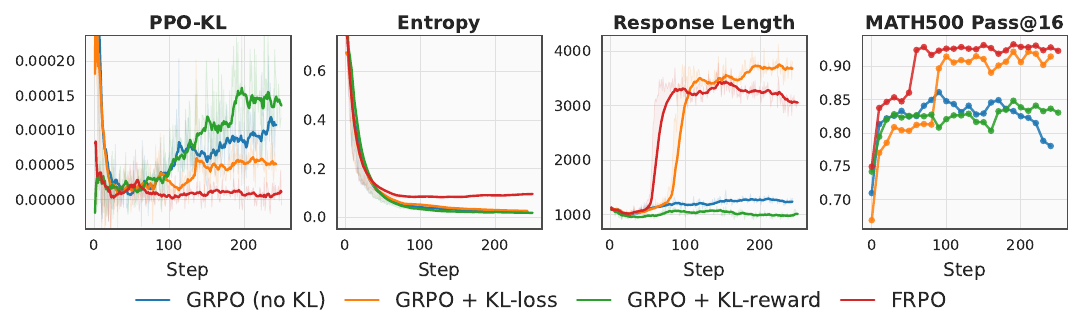}
    \caption{Training dynamics for Qwen3-4B-Base model with different KL integrations.}
    \label{fig:4b_full}
\end{figure}

The training dynamics of the Qwen3-4B-Base model are shown in Figure~\ref{fig:4b_full}. The training trend is similar to that of other models like Qwen3-1.7B-Base and Qwen3-30B-A3B-Base.

\subsection{Other Evaluation Metrics}
In previous sections, we mainly focused on the metric pass@n, as we hope that KL regularization could help maintain the base model's inherent ability and alleviate entropy collapse. From Figure~\ref{fig:main_30ba3b_reversekl} and Figure~\ref{fig:4b_full}, we could see that the convergence entropy of \method{} is indeed at a higher value, leading to a higher pass@n performance.
Here, we also show the mean@n performance.

\begin{figure}[h]
    \centering
    \includegraphics[width=0.55\linewidth]{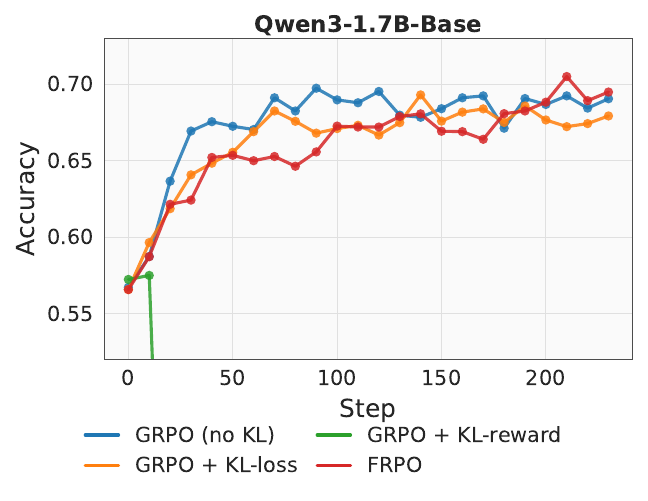}
    \caption{The training dynamics of mean@16 accuracy on MATH500 for Qwen3-1.7B-Base under different KL integrations.}
    \label{fig:mean_16_1.7b}
\end{figure}

\begin{figure}[h]
    \centering
    \includegraphics[width=0.55\linewidth]{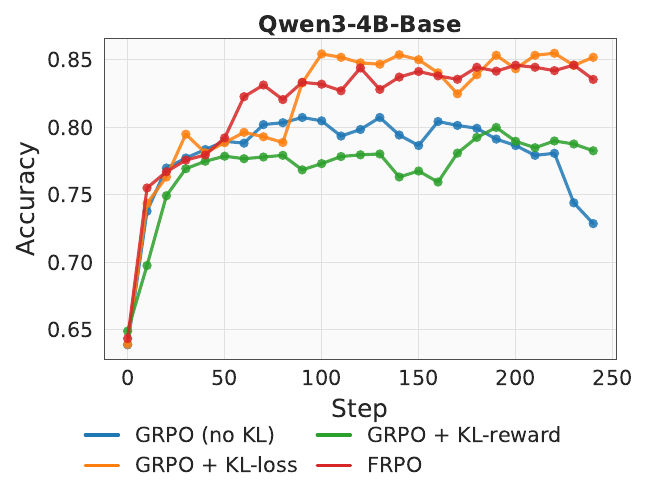}
    \caption{The training dynamics of mean@16 accuracy on MATH500 for Qwen3-4B-Base under different KL integrations.}
    \label{fig:mean_16_4b}
\end{figure}

\begin{figure}[h]
    \centering
    \includegraphics[width=0.5\linewidth]{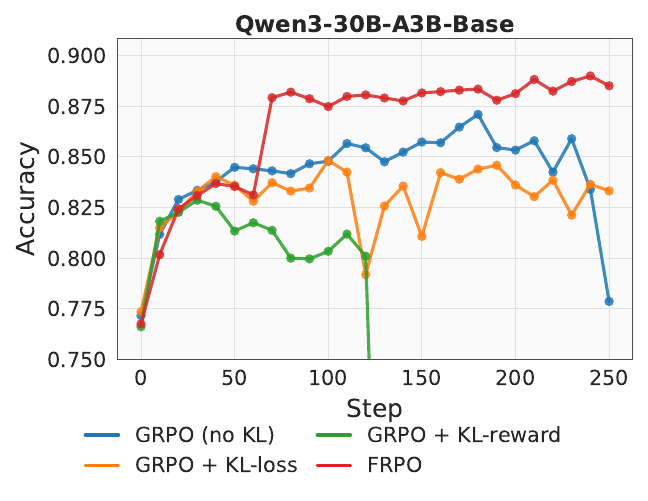}
    \caption{The training dynamics of mean@16 accuracy on MATH500 for Qwen3-30B-A3B-Base under different KL integrations.}
    \label{fig:mean_16_30ba3b}
\end{figure}

From Figure~\ref{fig:mean_16_1.7b}, Figure~\ref{fig:mean_16_4b}, and Figure~\ref{fig:mean_16_30ba3b}, we can see that \method{} not only improves pass@n performance but also achieves a comparable or even better mean@n accuracy than other methods that directly put KL in the loss or the reward. Especially on Qwen3-30B-A3B-Base, where off-policy issues are more pronounced than in other dense models due to its MoE structure, using future KL in \method{} helps stabilize training and yields a smoother training process.

\subsection{Full Evaluation Results}
We use AIME24, AIME25, AMC23, OlympiadBench \citep{he2024olympiadbench}, and MinervaMath \citep{lewkowycz2022solving} for evaluation besides MATH500. The information about each benchmark can be found in Table~\ref{tab:benchmark_info}

\begin{table}[h]
    \centering
    \caption{Information about benchmarks.}
    \label{tab:benchmark_info}
    \begin{tabular}{lc}
    \toprule
         \textbf{Benchmark} & \textbf{Size} \\\midrule
        AIME24 & 30 \\
        AIME25 & 30 \\
        AMC23 & 40 \\
        OlympiadBench & 674 \\
        MinervaMath & 272\\
        \bottomrule
    \end{tabular}
\end{table}

Table~\ref{tab:full_results_pass16} and Table~\ref{tab:full_results_mean16} summarize the evaluation results of all models with different KL integrations on several benchmarks for pass@16 and mean@16, respectively. Note that using (reverse) KL in the reward directly leads to model collapse.

\begin{table}[h]
    \centering
    \caption{Full evaluation results of mean@16 (\%) for Qwen3-1.7-Base, Qwen3-4B-Base, and Qwen3-30B-A3B-Base on different benchmarks.}
    \label{tab:full_results_mean16}
    \resizebox{\textwidth}{!}{
    \begin{tabular}{lcccccc}
    \toprule
        \textbf{Model} & \textbf{MATH500} & \textbf{AIME24} & \textbf{AIME25} & \textbf{AMC23} & \textbf{OlympiadBench} & \textbf{MinervaMath} \\\midrule
        Qwen3-1.7B-Base & 56.71 & 2.08 & 0.83 & 13.44 & 11.21 & 9.79 \\
        \quad + GRPO (no KL) & 69.23 & 8.33 & 4.17 & 40.16 & \textbf{33.25} & \textbf{28.33} \\
        \quad + GRPO KL-loss & 69.29 & 9.38 & 4.58 & \textbf{42.03} & 30.80 & 27.23 \\
        \quad + GRPO KL-reward & / & / & / & / & / & / \\
        \quad + \method{} & \textbf{70.90} & \textbf{9.79} & \textbf{4.79} & 41.56 & 31.64 & 25.46 \\\midrule
        Qwen3-4B-Base & 63.90 & 4.79 & 4.17 & 21.25 & 18.03 & 17.10 \\
        \quad + GRPO (no KL) & 80.41 & 16.25 & 12.29 & 61.41 & 44.95 & 39.48 \\
        \quad + GRPO KL-loss & \textbf{85.41} & \textbf{23.54} & \textbf{23.13} & \textbf{65.94} & \textbf{53.46} & \textbf{40.65} \\
        \quad + GRPO KL-reward & / & / & / & / & / & / \\
        \quad + \method{} & 84.11 & 21.67 & 19.58 & 63.28 & 51.03 & 39.22 \\\midrule
        Qwen3-30B-A3B-Base & 77.16 & 9.17 & 3.96 & 33.28 & 23.44 & 18.45\\
        \quad + GRPO (no KL) & 87.10 & 24.17 & 17.71 & 75.16 & 51.00 & \textbf{44.28} \\
        \quad + GRPO KL-loss & 84.83 & 22.29 & 10.21 & 66.56 & 47.53 & 42.69 \\
        \quad + GRPO KL-reward & / & / & / & / & / & / \\
        \quad + \method{} & \textbf{88.13} & \textbf{32.08} & \textbf{24.17} & \textbf{75.63} & \textbf{58.32} & 43.18 \\
        \bottomrule
    \end{tabular}}
\end{table}

\subsection{Using Current KL versus Stale KL}
To fully utilize the computation resources for the rollout engine, the rollout train batch is usually larger than the actor update mini-batch. In our experiments, the train batch size is $256$ prompts and the PPO mini-batch size is $64$ prompts. Thus, a rollout stage samples responses for $256$ prompts at once, and the actor then performs updates on four mini-batches. The PPO epoch is set to $1$, so each mini-batch is used once rather than repeatedly optimized. This design improves rollout throughput, while any staleness comes from policy changes across mini-batch updates within the same rollout batch; using multiple PPO epochs would introduce an additional stale-rollout effect.

\clearpage
\section{Case Studies}
\label{sec:app:case_study}

In this section, we briefly show concrete cases where models finetuned with the future-KL formulation answer hard problems that the model constrained with only the current single-step KL does not solve. From \ref{30ba3b_aime25}, we observe that the model finetuned with future KL typically induces longer responses, suggesting that \method{} can elicit long \gls{cot} behavior from the base model while maintaining the reference-policy constraint.

\begin{mybox}[30ba3b_aime25]{Qwen3-30B-A3B-Base AIME25}
    \textbf{Problem}

    Find the sum of all integer bases $b>9$ for which $17_b$ is a divisor of $97_b.$ Please reason step by step, and put your final answer within \verb|\boxed{}|.

    \textbf{Answer} 70.
    
    \textbf{\textcolor{red}{Wrong answer 1}}

    To solve the problem, we need to convert the numbers \(17_b\) and \(97_b\) from base \(b\) to base \(10\) and then find the integer bases \(b > 9\) for which \(17_b\) is a divisor of \(97_b\).

First, let's convert \(17_b\) and \(97_b\) to base \(10\):
\[
17_b = 1 \cdot b + 7 = b + 7
\]
\[
97_b = 9 \cdot b + 7 = 9b + 7
\]

We need to find the integer bases \(b > 9\) such that \(b + 7\) is a divisor of \(9b + 7\). This means that there exists an integer \(k\) such that:
\[
9b + 7 = k(b + 7)
\]

Rearranging the equation, we get:
\[
9b + 7 = kb + 7k
\]
\[
9b - kb = 7k - 7
\]
\[
b(9 - k) = 7(k - 1)
\]
\[
b = \frac{7(k - 1)}{9 - k}
\]

Since \(b\) is an integer, \(\frac{7(k - 1)}{9 - k}\) must be an integer. Therefore, \(9 - k\) must be a divisor of \(7(k - 1)\). Let's find the possible values of \(k\).

The divisors of \(7\) are \(\pm 1, \pm 7\). We will check each case:

\begin{enumerate}
    \item \(9 - k = 1 \implies k = 8\)
    \[
    b = \frac{7(8 - 1)}{1} = 49
    \]

    \item \(9 - k = -1 \implies k = 10\)
    \[
    b = \frac{7(10 - 1)}{-1} = -63
    \]
    not valid since \(b > 9\).

    \item \(9 - k = 7 \implies k = 2\)
    \[
    b = \frac{7(2 - 1)}{7} = 1
    \]
    not valid since \(b > 9\).

    \item \(9 - k = -7 \implies k = 16\)
    \[
    b = \frac{7(16 - 1)}{-7} = -15
    \]
    not valid since \(b > 9\).
\end{enumerate}

The only valid value is \(b = 49\). We need to check if there are any other possible values of \(k\). Since \(9 - k\) must be a divisor of \(7\), we have checked all possible divisors.

Therefore, the only integer base \(b > 9\) for which \(17_b\) is a divisor of \(97_b\) is \(b = 49\). The sum of all such integer bases is:
\[
\boxed{49}
\]
    
    \textbf{\textcolor{red}{Wrong answer 2}}

    To solve the problem, we need to interpret the numbers \(17_b\) and \(97_b\) in base \(b\) and determine the integer bases \(b > 9\) for which \(17_b\) is a divisor of \(97_b\).

First, let's convert \(17_b\) and \(97_b\) from base \(b\) to base \(10\):
\begin{itemize}
    \item \(17_b = 1 \cdot b + 7 = b + 7\)
    \item \(97_b = 9 \cdot b + 7 = 9b + 7\)
\end{itemize}

We need to find the integer bases \(b > 9\) such that \(b + 7\) is a divisor of \(9b + 7\). This means that there exists an integer \(k\) such that:
\[
9b + 7 = k(b + 7)
\]

Rearranging the equation, we get:
\[
9b + 7 = kb + 7k
\]
\[
9b - kb = 7k - 7
\]
\[
b(9 - k) = 7(k - 1)
\]
\[
b = \frac{7(k - 1)}{9 - k}
\]

Since \(b\) is an integer, \(\frac{7(k - 1)}{9 - k}\) must be an integer. Therefore, \(9 - k\) must be a divisor of \(7(k - 1)\). Let's find the possible values of \(k\) such that \(b > 9\).

We will iterate over possible values of \(k\) and check if \(b\) is an integer greater than \(9\).

Let's write the code to find the valid bases \(b\).

\begin{verbatim}
# Function to find the sum of all valid bases b > 9
def find_valid_bases():
    valid_bases = []
    # Iterate over possible values of k
    for k in range(1, 9):
        if (9 - k) != 0 and (7 * (k - 1)) % (9 - k) == 0:
            b = (7 * (k - 1)) // (9 - k)
            if b > 9:
                valid_bases.append(b)
    return valid_bases

# Find the valid bases
valid_bases = find_valid_bases()
# Calculate the sum of valid bases
sum_valid_bases = sum(valid_bases)
print((valid_bases, sum_valid_bases))
\end{verbatim}

\begin{verbatim}
([32, 57], 89)
\end{verbatim}

The valid bases \(b > 9\) for which \(17_b\) is a divisor of \(97_b\) are \(32\) and \(57\). The sum of these bases is \(32 + 57 = 89\).

Therefore, the final answer is \(\boxed{89}\).
    
    \textbf{\textcolor{OliveGreen}{Correct answer 1}}

    Alright, I have this problem:

\[
\textbf{Find the sum of all integer bases } b > 9 \textbf{ for which } 17_b \textbf{ is a divisor of } 97_b.
\]

First, I need to understand what \(17_b\) and \(97_b\) mean. The subscript \(b\) indicates that these numbers are in base \(b\). So, \(17_b\) means \(1 \times b + 7\) in base \(10\), and \(97_b\) means \(9 \times b + 7\) in base \(10\).

Let's write that down:

\[
17_b = 1 \times b + 7 = b + 7
\]
\[
97_b = 9 \times b + 7 = 9b + 7
\]

The problem states that \(17_b\) is a divisor of \(97_b\). In other words, \(b + 7\) divides \(9b + 7\) without leaving a remainder. Mathematically, this can be expressed as:

\[
9b + 7 \equiv 0 \pmod{b + 7}
\]

This means that \(9b + 7\) is a multiple of \(b + 7\). To find the values of \(b\) that satisfy this condition, I need to solve the congruence.

Let's denote \(d = b + 7\). Then, \(b = d - 7\). Substitute \(b\) in the expression \(9b + 7\):

\[
9b + 7 = 9(d - 7) + 7 = 9d - 63 + 7 = 9d - 56
\]

So, the congruence becomes:

\[
9d - 56 \equiv 0 \pmod{d}
\]

Simplifying this, we get:

\[
9d \equiv 56 \pmod{d}
\]

But \(9d\) is a multiple of \(d\), so \(9d \equiv 0 \pmod{d}\). Therefore:

\[
0 \equiv 56 \pmod{d}
\]

This implies that \(d\) divides \(56\). So, \(d\) is a divisor of \(56\).

Let's list all the positive divisors of \(56\):

\[
1, 2, 4, 7, 8, 14, 28, 56
\]

But we know that \(d = b + 7\) and \(b > 9\). Therefore:

\[
d = b + 7 > 9 + 7 = 16
\]

So, \(d\) must be a divisor of \(56\) that is greater than \(16\). Looking at the list of divisors:

\[
28, 56
\]

So, \(d\) can be \(28\) or \(56\).

Now, let's find the corresponding \(b\) for each \(d\):

\begin{enumerate}
    \item If \(d = 28\):
    \[
    b = d - 7 = 28 - 7 = 21
    \]

    \item If \(d = 56\):
    \[
    b = d - 7 = 56 - 7 = 49
    \]
\end{enumerate}

So, the possible values of \(b\) are \(21\) and \(49\).

Finally, the problem asks for the sum of all such \(b\):

\[
21 + 49 = 70
\]

Therefore, the sum of all integer bases \(b > 9\) for which \(17_b\) is a divisor of \(97_b\) is:

\[
\boxed{70}
\]

\textbf{\textcolor{OliveGreen}{Correct answer 2}}

Alright, I have this problem:

\[
\textbf{Find the sum of all integer bases } b > 9 \textbf{ for which } 17_b \textbf{ is a divisor of } 97_b.
\]

First, I need to understand what \(17_b\) and \(97_b\) mean. The subscript \(b\) indicates that these numbers are in base \(b\). So, \(17_b\) and \(97_b\) are numbers written in base \(b\).

\subsection*{Step 1: Convert \(17_b\) and \(97_b\) to Base 10}

To work with these numbers more easily, I'll convert them to base \(10\).

\begin{enumerate}
    \item \textbf{Converting \(17_b\) to base \(10\):}

    In base \(b\), the number \(17_b\) represents:
    \[
    1 \times b^1 + 7 \times b^0 = b + 7
    \]

    \item \textbf{Converting \(97_b\) to base \(10\):}

    Similarly, \(97_b\) represents:
    \[
    9 \times b^1 + 7 \times b^0 = 9b + 7
    \]
\end{enumerate}

So, in base \(10\), \(17_b\) is \(b + 7\) and \(97_b\) is \(9b + 7\).

\subsection*{Step 2: Understand the Divisibility Condition}

The problem states that \(17_b\) is a divisor of \(97_b\). In base \(10\) terms, this means that \(b + 7\) divides \(9b + 7\) without leaving a remainder.

Mathematically, this can be expressed as:
\[
9b + 7 \equiv 0 \pmod{b + 7}
\]
or
\[
b + 7 \mid 9b + 7
\]

\subsection*{Step 3: Simplify the Divisibility Condition}

To find the values of \(b\) that satisfy this condition, I'll use the concept of modular arithmetic.

We have:
\[
9b + 7 \equiv 0 \pmod{b + 7}
\]

This means that \(9b + 7\) is a multiple of \(b + 7\). Let's express \(9b + 7\) in terms of \(b + 7\):

\[
9b + 7 = 9(b + 7) - 63 + 7 = 9(b + 7) - 56
\]

So,
\[
9b + 7 = 9(b + 7) - 56
\]

Since \(b + 7\) divides \(9b + 7\), it must also divide the remainder when \(9b + 7\) is divided by \(b + 7\). In this case, the remainder is \(-56\), but since we're dealing with divisibility, we can consider the absolute value:

\[
b + 7 \mid 56
\]

\subsection*{Step 4: Find All Divisors of 56}

Now, I need to find all integers \(b + 7\) that divide \(56\). First, let's list all the positive divisors of \(56\):

\[
1, 2, 4, 7, 8, 14, 28, 56
\]

Since \(b > 9\), \(b + 7 > 16\). So, we'll only consider divisors of \(56\) that are greater than \(16\).

Looking at the list:

\begin{itemize}
    \item \(1: \leq 16\) \(\to\) Skip
    \item \(2: \leq 16\) \(\to\) Skip
    \item \(4: \leq 16\) \(\to\) Skip
    \item \(7: \leq 16\) \(\to\) Skip
    \item \(8: \leq 16\) \(\to\) Skip
    \item \(14: \leq 16\) \(\to\) Skip
    \item \(28: > 16\) \(\to\) Keep
    \item \(56: > 16\) \(\to\) Keep
\end{itemize}

So, the possible values for \(b + 7\) are \(28\) and \(56\).

\subsection*{Step 5: Solve for \(b\)}

Now, let's find the corresponding values of \(b\):

\begin{enumerate}
    \item \textbf{Case 1:} \(b + 7 = 28\)
    \[
    b = 28 - 7 = 21
    \]

    \item \textbf{Case 2:} \(b + 7 = 56\)
    \[
    b = 56 - 7 = 49
    \]
\end{enumerate}

\subsection*{Step 6: Verify the Solutions}

Before finalizing, I should verify that these values of \(b\) satisfy the original condition.

\begin{enumerate}
    \item \textbf{For \(b = 21\):}
    \begin{itemize}
        \item \(17_{21} = 21 + 7 = 28\)
        \item \(97_{21} = 9 \times 21 + 7 = 189 + 7 = 196\)
        \item Check if \(28\) divides \(196\):
        \[
        196 \div 28 = 7
        \]
        Yes, it does.
    \end{itemize}

    \item \textbf{For \(b = 49\):}
    \begin{itemize}
        \item \(17_{49} = 49 + 7 = 56\)
        \item \(97_{49} = 9 \times 49 + 7 = 441 + 7 = 448\)
        \item Check if \(56\) divides \(448\):
        \[
        448 \div 56 = 8
        \]
        Yes, it does.
    \end{itemize}
\end{enumerate}

Both values of \(b\) satisfy the condition.

\subsection*{Step 7: Calculate the Sum of All Valid \(b\)}

The problem asks for the sum of all integer bases \(b > 9\) that satisfy the condition. We found two such values: \(21\) and \(49\).

\[
21 + 49 = 70
\]

\subsection*{Final Answer}

\[
\boxed{70}
\]
\end{mybox}
\section{License of Assets}
\label{sec:app:license}

\begin{table}[h]
    \centering
    \caption{License of assets used in this work.}
    \label{tab:license}
    \begin{tabular}{lc}
    \toprule
        \textbf{Asset} & \textbf{License} \\\midrule
         Qwen3 models & apache-2.0 \\
         DAPO-Math-17k dataset & apache-2.0 \\
         code framework (verl) & apache-2.0 \\
         \bottomrule
    \end{tabular}
\end{table}
\section{Limitations}
\label{sec:app:limitations}

Our theoretical analysis identifies the future-KL term in the on-policy gradient of a specified KL-regularized GRPO-style objective, and our empirical results suggest that this credit-assignment signal can improve reasoning RL in the tested setting. The conclusion may need to be more generally verified on different model architectures, scales, datasets, and KL estimator choices. Due to the computation resources budget, we only conduct experiments on the Qwen series base models with a few scales ourselves. Besides, staleness or off-policy effects from other sources, such as training-inference mismatch \citep{yao2025offpolicy,qi2025defeating}, could also lead to training instability. The mutual interactions among all these factors are still underexplored. We leave these valuable directions for possible future research. 

\end{document}